\documentclass[10pt,twocolumn,letterpaper]{article}

\usepackage[pagenumbers]{iccv} 

\newcommand{\HL}[1]{\textbf{\cellcolor{blue!15}#1}}
\newcommand{\HB}[1]{\cellcolor{blue!15}#1}

\usepackage{multirow}
\usepackage{makecell}
\usepackage{soul}

\usepackage[utf8]{inputenc}
\usepackage{indentfirst}
\usepackage{comment}
\usepackage{graphicx}
\usepackage[a4paper, total={6in, 8in}]{geometry}
\usepackage{amssymb}
\usepackage{pifont}

\usepackage{mathtools}

\usepackage{amsmath}
\DeclareMathOperator*{\argmax}{arg\,max}

\usepackage{listings}

\usepackage{afterpage}

\usepackage{array}
\newcolumntype{T}{>{\tiny}l}
\newcolumntype{S}{>{\scriptsize}c}

\definecolor{iccvblue}{rgb}{0.21,0.49,0.74}
\usepackage[pagebackref,breaklinks,colorlinks,allcolors=iccvblue]{hyperref}

\def\paperID{9116} 
\def\confName{ICCV}
\def\confYear{2025}

\title{Making Better Mistakes in CLIP-Based Zero-Shot Classification \\with Hierarchy-Aware Language Prompts}

\author{Tong Liang\\
Ohio State University\\
Columbus, Ohio, USA\\
{\tt\small liang.693@osu.edu}
\and
Jim Davis\\
Ohio State University\\
Columbus, Ohio, USA\\
{\tt\small davis.1719@osu.edu}
}

\begin{document}
\maketitle
\begin{abstract}

Recent studies are leveraging advancements in large language models (LLMs) trained on extensive internet-crawled text data to generate textual descriptions of downstream classes in CLIP-based zero-shot image classification. While most of these approaches aim at improving accuracy, our work focuses on ``making better mistakes", of which the mistakes' severities are derived from the given label hierarchy of downstream tasks. Since CLIP's image encoder is trained with language supervising signals, it implicitly captures the hierarchical semantic relationships between different classes. This motivates our goal of making better mistakes in zero-shot classification, a task for which CLIP is naturally well-suited. Our approach (HAPrompts) queries the language model to produce textual representations for given classes as zero-shot classifiers of CLIP to perform image classification on downstream tasks. To our knowledge, this is the first work to introduce making better mistakes in CLIP-based zero-shot classification. Our approach outperforms the related methods in a holistic comparison across five datasets of varying scales with label hierarchies of different heights in our experiments. Our code and LLM-generated image prompts: \href{https://github.com/ltong1130ztr/HAPrompts}{https://github.com/ltong1130ztr/HAPrompts}.
\end{abstract}    
\section{Introduction} \label{sec:intro}

\begin{figure}
    \centering
    \includegraphics[height=1.6in]{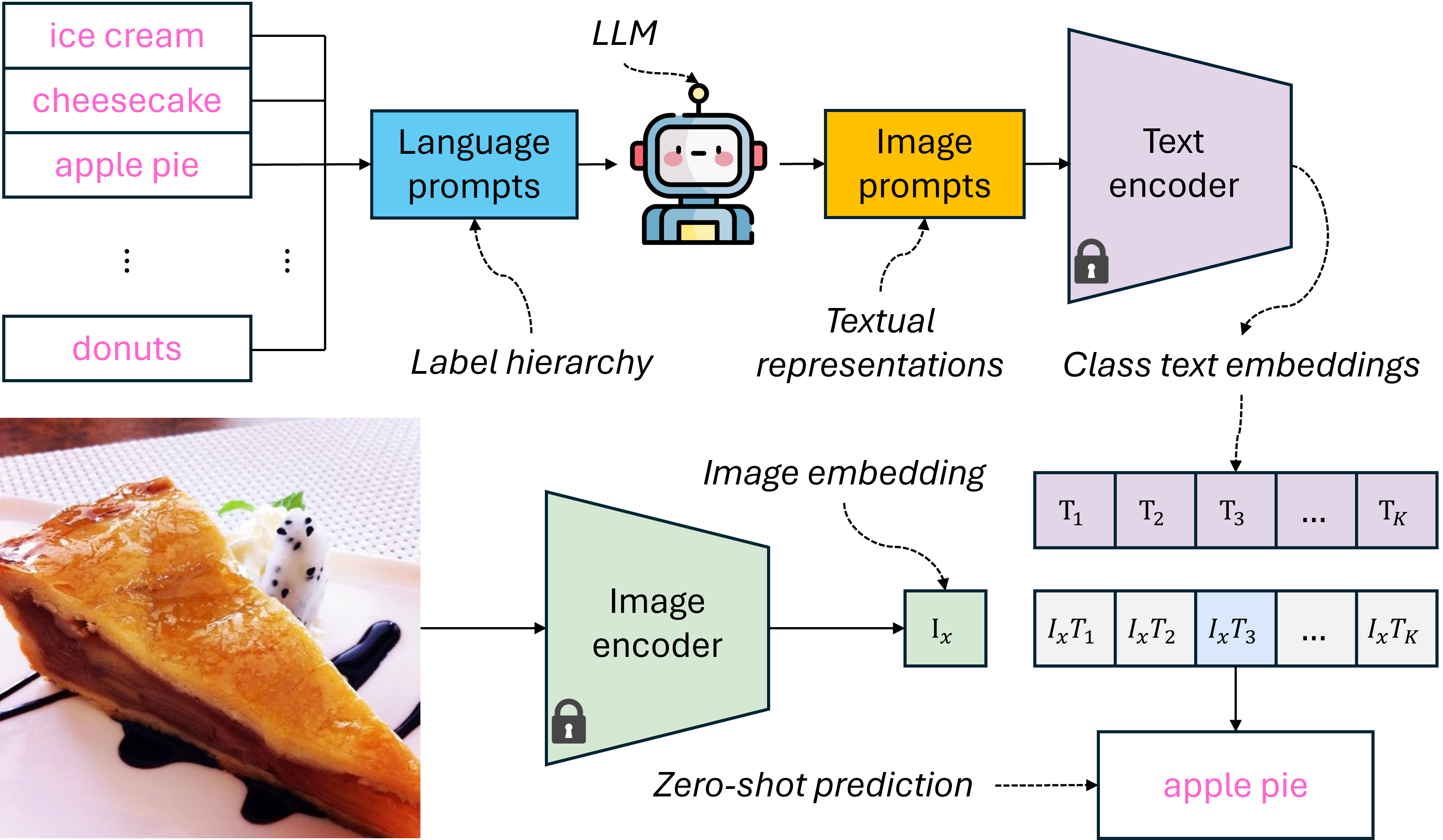}
    \caption{Overall process of the proposed CLIP-based zero-shot classification. We introduce prior knowledge of the label hierarchy to the language prompts used to query the LLM.}
    \label{fig:summary}
\end{figure}

\begin{figure*}[t]
    \centering 
    \includegraphics[height=1.7in]{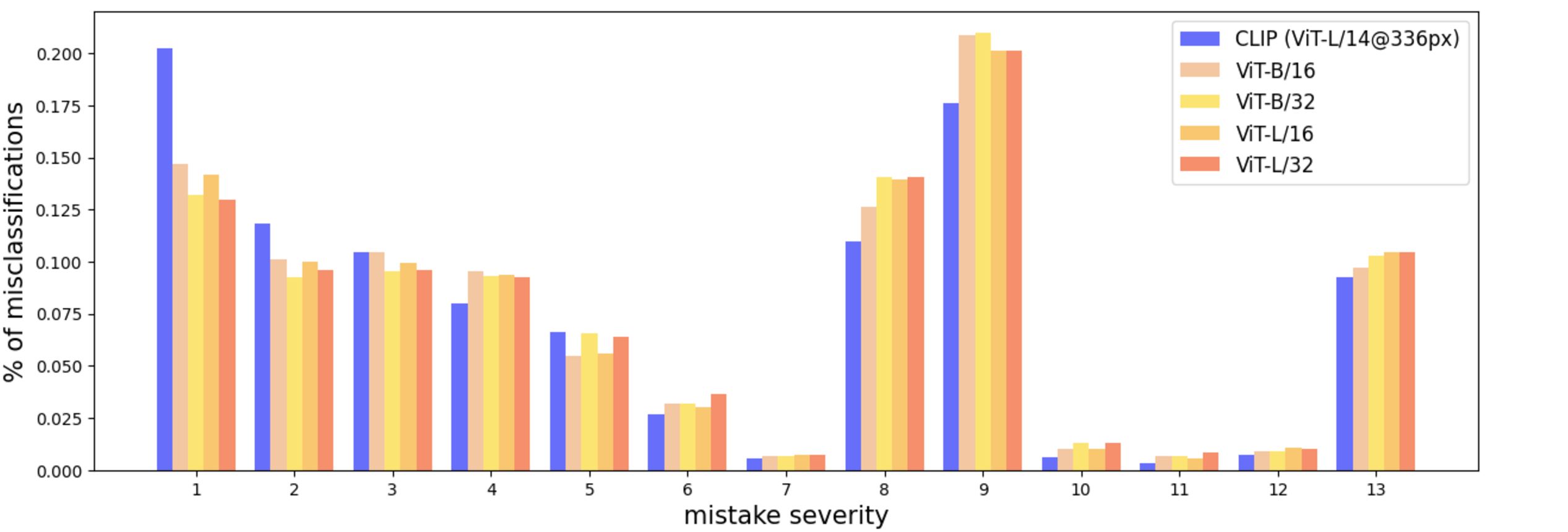} 
    \caption{Histogram of mistake severities for predictions of CLIP (zero-shot) and ViTs (trained with ImageNet from scratch) on ImageNet. The mistake severities of predictions are derived from our label hierarchy of ImageNet. ViTs trained from scratch tend to make more high-severity mistakes (severity $\ge$ 7 in ImageNet). The ViT models are acquired from PyTorch pretrained weights \cite{ViT_weights}.}
    \label{fig:clip-vs-vit}
\end{figure*}

The idea of leveraging language information to enhance the learning of visual features is primarily popularized by the studies on vision-language pretraining, such as CLIP \cite{CLIP} and ALIGN \cite{ALIGN}. In \cite{CLIP, ALIGN, MERU}, the image-text joint embedding is learned with paired image-text data where the text data are natural language text descriptions of the images. UniCL \cite{UniCL} proposes a framework to unify contrastive learning with the image-text and the image-label data. K-LITE \cite{K_lite} leverages structured external knowledge from WordNet \cite{WordNet} and Wiktionary \cite{Wiktionary} to augment the original text descriptions paired with the image data. BioCLIP \cite{BioClip} uses ordered sequences of labels derived from biological taxonomy or common class names instead of text phrases to pretrain a vision language model (VLM) grounded in biology. GLIP \cite{GLIP} learns to pair image regions with the token embeddings of the paired text descriptions to achieve object detection. LAVILA \cite{NarratorRephraser_VLM} uses a pretrained image-to-text model to output densely annotated text descriptions for video data for subsequent video representation learning. Flamingo \cite{flamingo_VLM} connects frozen image and text encoders with trainable cross-attention modules to learn visual representations. 
The above works mainly focus on developing foundational VLMs and only BioCLIP and K-LITE explicitly employ hierarchical label relationships for training. 
Our approach focuses on downstream tasks of VLM, particularly image classification using CLIP. 
We aim to leverage LLM to enhance the semantic relationships between downstream classes. This enables us to make predictions with reduced mistake severity in zero-shot transfer learning. Specifically, we incorporate label hierarchies structured as trees (where each node has at most one parent) to perform CLIP-based zero-shot classification at the leaf level.

We prompt the LLM to generate textual representations for the given classes of the downstream task and employ CLIP for zero-shot classification based on the LLM-generated text description of the classes. The overall zero-shot classification process is summarized in Fig.~\ref{fig:summary}. Our approach does not treat all mistakes equally, as some mistakes are more severe than others in real-world applications, e.g., mistaking a human as a pole in autonomous driving. 
Following \cite{BetterMistakes_hie_loss}, we measure mistake severity with the hierarchical distance between the incorrect prediction and its ground truth. We propose to generate textual descriptions of the given classes based on the label hierarchy of the downstream dataset to reduce mistake severities of zero-shot predictions. CLIP pretrained with image-text pairs is well suited for the task as it uses natural language text data for supervision to train its image encoder. The resulting image-text joint embedding space implicitly captures the hierarchical semantic similarity between classes. In Fig.~\ref{fig:clip-vs-vit}, we show that CLIP trained with image-text contrastive loss is inherently prone to make less severe mistakes than Vision Transformers (ViTs) \cite{ViT_paper} trained from scratch on ImageNet-1k \cite{Imagenet_1k} with one-hot labels and cross-entropy loss. The difference in error structures suggests that CLIP is better at capturing the semantic similarity between different classes in natural language image captions. This motivates us to take advantage of CLIP to enhance its knowledge of hierarchical label relationships further and reduce the severities of mistakes in its predictions. There are several contributions of our work:
\begin{itemize}
    \item Our work first introduces ``making better mistakes" as a suitable goal for CLIP-based zero-shot transfer learning.
    \item Our approach is training-free. We only query LLM to generate image prompts for subsequent zero-shot classification.
    \item Our language prompts are dataset-agnostic. We employ the same approach to construct language prompts for classes of a given dataset and its hierarchy.
    \item Our approach outperforms related methods in a holistic comparison across five datasets.
\end{itemize}

\section{Related Works}
\label{sec:related_works}
\noindent\textbf{CLIP-based zero-shot classification:} This section mainly focuses on methods adopting CLIP as their foundational VLM for zero-shot transfer learning to downstream tasks. The text encoder of CLIP serves as an interface to receive extra structural/semantic information about the target classes provided by the downstream tasks. The extra information provided by the task or acquired from external sources, such as a knowledge base or LLM, can enhance the visual recognition performance of CLIP in zero-shot settings. In CLIP, the authors employ a set of $M$ text templates manually crafted for each dataset. They are dataset-specific, class-agnostic, and shared across all classes to form a set of textual representations per class. For example, \textcolor{orange}{``a photo of a \{classname\}''}, and  \textcolor{orange}{``i love my \{classname\}!''}\footnote{We highlight image prompts used to generate class text embeddings for CLIP with the \textcolor{orange}{orange} color to distinguish such prompts from LLM-querying language prompts highlighted with \textcolor{blue}{blue} color.} are two of the context-prompts used for StandfordCars \cite{standfordcars} in CLIP. When embedded with the real class name, we refer to these context-prompts as \textit{image prompts} for the class. The image prompts for a given class are passed to the text encoder to acquire the respective text embeddings, which are then aggregated into a single text embedding per class. 
The resulting text embedding for each class serves as its zero-shot classifier. To boost CLIP's zero-shot capability, one could enhance the textual representations of target classes with more descriptive information.
Recent studies \cite{CHiLS, CuPL, VCD_LLM, HieGroupComp} leverage the power of LLMs to alleviate the effort of manually crafting class-specific image prompts.
These works query an LLM to generate class-specific textual representations, replacing the hand-crafted dataset-specific image prompts.

In CHiLS\cite{CHiLS}, GPT-3 \cite{GPT3} is used to generate a fixed number of possible child classes for each real class label in the dataset. The text embeddings of real (coarse level) classes and LLM-generated (fine-grained level) pseudo child classes are employed to produce softmax predictions in their respective levels. The authors use coarse-level softmax to reweight the fine-grained level softmax following their hierarchical relationships to reconcile inconsistent predictions between the coarse and fine-grained levels. 
Compared with our approach, CHiLS only leverages one level of parent-child hierarchical label relationships and assumes the hierarchy is balanced with a common branching factor. This assumption does not hold well. The hierarchies of the datasets \cite{SUN397, fgvc_aircraft, iNat2019, cub200, Imagenet_1k} are often skewed, with some coarse classes having more children than others. 

In CuPL \cite{CuPL}, the authors propose to design three to five dataset-specific \textit{language prompts} to query GPT-3 about the classes of downstream datasets. For example, the language prompts employed for the CUB-200 \cite{cub200} bird dataset are: \textcolor{blue}{``Describe what the bird \{classname\} looks like.''}, \textcolor{blue}{``Describe the bird \{classname\}.''}, and \textcolor{blue}{``What are the identifying characteristics of the bird \{classname\}?''}. The resulting LLM response is the textual representation of the associated class. This approach greatly reduces the manual effort required to curate image prompts by querying LLM (e.g., the number of manually crafted image prompts used in CLIP ranges from one for Food-101 \cite{food101} to 80 for ImageNet-1k \cite{Imagenet_1k}).

In VCD \cite{VCD_LLM}, the LLM-prompting technique is also employed to query for text descriptions of target classes. 
However, VCD aggregates information provided by the textual representations of a class in the log probability space instead of the text embedding space like CLIP and CuPL. To encourage GPT-3 to produce a list of descriptors for a query class instead of a general summary, VCD uses a question-answer pair as the language prompt: \textcolor{blue}{``Q: What are useful features for distinguishing a \{classname\} in a photo?''}, \textcolor{blue}{``A: There are several useful visual features to tell there is a \{classname\} in a photo: -''}. VCD predictions are naturally explainable, as one can examine the descriptor-based similarity scores of a given class to find the primary contributors of its predictions and the respective descriptions. 

In HIE \cite{HieGroupComp}, a group-and-compare iterative approach is proposed to learn a tree of the classes based on their text embeddings derived from LLM-generated descriptions. The authors initialize the class embeddings using textual representations of each class generated by ChatGPT \cite{chatGPT} following a similar language prompt in VCD \cite{VCD_LLM}. The classes are then grouped with k-means clustering using the initial class embeddings. For each class, its textual representation is updated by prompting the language model to compare this class with the rest of the classes clustered in the same group using a question-answer pair: \textcolor{blue}{``Q: What are useful features for distinguishing a \{classname\} from \{group description\} in a photo?''}, \textcolor{blue}{``A: -''}. Depending on the group size, the prompt replaces the above \textcolor{blue}{`group description'} with either a one-sentence LLM-generated summary of the group or a list of the other classes in the group. 
The resulting class hierarchy and the intermediate textual representations of each class are leveraged to compute a multi-level fused score for zero-shot inference. 

\noindent\textbf{Making better mistakes.} 
This section mainly introduces works incorporating prior knowledge of label hierarchy to make better mistakes. 
In \cite{BetterMistakes_hie_loss}, a hierarchical cross-entropy loss is proposed to learn hierarchy-aware representations. The authors also proposed an alternative approach that minimizes the KL-divergence between softmax scores and soft-label embeddings derived from the label hierarchy. 
In \cite{flamingo_cub200_tree}, the authors employ multiple classification heads for different levels of classes in the hierarchy to train a ResNet \cite{ResNet} backbone for the respective coarse and fine-grained classifiers. 
The multi-classification heads approach is also adopted by \cite{HAFeature}, where the authors minimize the Jensen-Shannon Divergence \cite{JSDivergence} between coarse-level and fine-grained level predictions to enforce the hierarchical label relationships. 
In \cite{HAFrame}, a pre-computed hierarchy-aware frame is frozen as the linear classifiers to induce a neural collapse \cite{neuralcollapse} of the penultimate features onto the frame during training for reducing the mistake severity.
In \cite{H_ensemble}, an ensemble of two models trained separately with coarse-grained and fine-grained labels is proposed to modify fine-grained predictions with coarse-grained softmax scores at test time. 
In CRM \cite{BetterMistakes_CRM}, the conditional risk of predictions derived from the label hierarchy is employed to amend the backbone network predictions at test time. 
\section{Method}
In this section, we introduce the preliminaries of CLIP-based zero-shot classification and our proposed language prompts to generate hierarchy-aware textual representations, i.e., image prompts of the downstream classes. 
\subsection{Preliminaries}\label{sect:preliminaries}
Our approach adopts the pretrained image and text encoders from CLIP and keeps them frozen during inference. For a given downstream dataset and its label hierarchy, we query an LLM about the leaf classes $\mathcal{Y}=\{y_1,...,y_K\}$ with hierarchy-aware \textcolor{blue}{\textit{language prompts}} $\mathcal{P}_{y_k}$, $y_k\in\mathcal{Y}$ to generate their respective \textcolor{orange}{\textit{image prompts}}, i.e., LLM-generated textual representations of class $y_k$.  
These image prompts are passed to the text encoder to acquire their associated text embeddings. Following the ensemble approach in CLIP, the text embeddings derived from image prompts of the class $y_k$ are L2-normalized, averaged, and L2-normalized again to form a single vector representation $T_k$ as the overall class text embedding for $y_k$. The class text embeddings $\{T_1,...,T_K\}$ are the respective classifiers for the subsequent zero-shot classification with image embeddings $I_x$ of input image $x$. As shown in Fig.~\ref{fig:summary}, we use the dot product between the input image embedding and class text embeddings to produce zero-shot prediction:
\begin{equation}
    \hat{y}_x := \argmax_k\: I_x\cdot T_k
\end{equation}

\subsection{Hierarchy-Aware Language Prompts} \label{sect:language_prompts}

\begin{figure}
    \centering
    \includegraphics[height=2.4in]{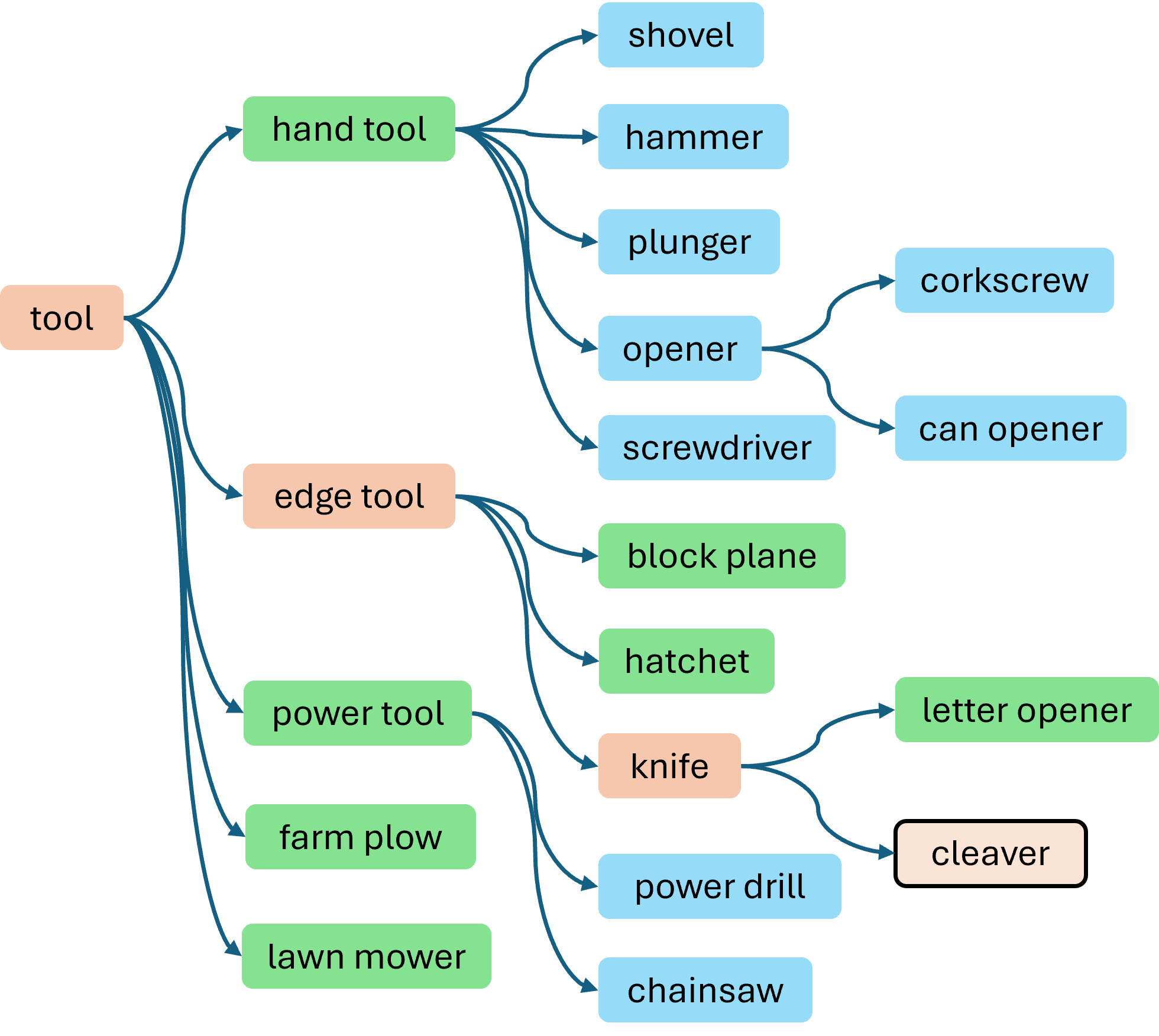}
    \caption{Example of a subtree of the ImageNet hierarchy employed in our approach. }
    \label{fig:tool_tree}
\end{figure}

We apply two types of language prompts for each leaf class of the downstream dataset: (1) comparative prompts and (2) path-based prompts. The comparative prompts strengthen the difference between the leaf class and its related classes in the hierarchy, whereas the path-based prompts focus on the unique characteristics of the leaf class with inserted hints of its ancestor classes in the hierarchy. For each leaf class in the query, the two groups of language prompts are constructed as input for LLM to generate their respective image prompts. 

\noindent\textbf{Comparative prompts.} Given a leaf class $y_k$, we compare it with its leaf-level peers and ancestor-level peers in the hierarchy. Both comparisons share the same template to construct their language prompts $\mathcal{P}^{LP}_{y_k}$ and $\mathcal{P}^{AP}_{y_k}$, respectively. The target leaf class $y_k$ is referred to as the \textit{query class}, and the associated leaf or ancestor-level peer classes are referred to as the \textit{related class}: 
{\color{blue}
\begin{lstlisting}[basicstyle=\ttfamily\small]
How does {query class} look 
differently from {related class}?
\end{lstlisting}
}
\noindent In Fig.~\ref{fig:tool_tree}, we give an example label hierarchy where the query class is \textit{`cleaver'}, and it has two ancestors (excluding the root \textit{`tool'}) \textit{`knife'} and \textit{`edge tool'} in the hierarchy. The associated leaf and ancestor-level peers for comparisons with \textit{`cleaver'} are highlighted in light green. For example, \textit{`letter opener'} is leaf-level peer of \textit{`cleaver'}, \textit{`hatchet'} is peer of \textit{`knife'}, an ancestor-level peer of \textit{`cleaver'}. 
The resulting image prompts stress the difference between the query and related classes along its ancestral path. 
Our approach reduces the exhaustive 1-vs-rest comparisons between the query class and all other leaf classes to comparisons between the query class and its leaf and ancestor-level peers. 
Therefore, leaf classes that are hierarchically close to each other are compared at a fine-grained level (e.g., \textit{`cleaver'} vs \textit{`letter opener'}). Hierarchically distant leaf classes are represented by their ancestor and compared at a coarse level (e.g., \textit{`cleaver'} vs \textit{`chainsaw'} and \textit{`cleaver'} vs \textit{`power drill'} are summarized to one comparison:  \textit{`cleaver'} vs \textit{`power tool'}).

\noindent\textbf{Path-based prompts.} Besides comparative prompts, we also use three generic prompts for all leaf classes incorporated with hints of their ancestor classes (excluding the root) in the label hierarchy to form our path-based language prompts:
{\color{blue}
\begin{lstlisting}[basicstyle=\ttfamily\small]
What does {query class} (a type of 
{ancestor class}) look like?

Describe a picture of {query class} 
(a type of {ancestor class}).

What are the unique characteristics 
of {query class} (a type of {ancestor 
class})?
\end{lstlisting}
}
\noindent Our path-based prompts complement the comparative prompts and emphasize the distinct features of the query class. We observed in our preliminary experiments that the added hint of the ancestor class also helps clarify the query class, as sometimes LLM confuses the query class with other classes (e.g., \textit{`hay'} as animal feed in ImageNet is mistaken as \textit{`hey'}, the greeting phrase). For a given leaf class $y_k$, we construct its path-based prompts $\mathcal{P}^G_{y_k}$ by iterating through its non-root ancestor classes and employing the above three templates for each leaf-ancestor class pair. For a given leaf class $y_k$, our approach feeds its language prompts $\mathcal{P}_{y_k}$:
\begin{equation}
\mathcal{P}_{y_k}:=\mathcal{P}^{LP}_{y_k}\cup \mathcal{P}^{AP}_{y_k} \cup \mathcal{P}^G_{y_k}    
\end{equation}
to an LLM to generate the respective image prompts. We aggregate the text embeddings of image prompts, as introduced in Sect.~\ref{sect:preliminaries}, to form the associated class text embedding for subsequent zero-shot classification.

\section{Experiments}

\begin{table}[b]
    \centering
    \scriptsize
    \begin{tabular}{l|c|c|c|c}
         \hline
         dataset & height & leaf classes &  val  & test \\
         \hline
         
         Food-101 \cite{food101} & 4 & 101 & 25250 & 25250 \\
         UCF-101 \cite{ucf101} & 3 &  101  & 9537 & 3783 \\
         CUB-200 \cite{cub200} & 3  & 200  & 4000 & 5794 \\
         SUN-324 \cite{SUN397} & 4 & 324  & 8100 & 16200 \\
         ImageNet-1k \cite{ILSVRC15} & 13 & 998 & 24950 & 24950\\
         \hline
         
    \end{tabular}
    \caption{Statistics of the datasets in our experiments.}
    \label{tab:datasets}
\end{table}

We compare the performance of our approach and related methods on five datasets: Food-101 \cite{food101}, UCF-101 \cite{ucf101}, CUB-200 \cite{cub200}, SUN-324 \cite{SUN397}, and ImageNet-1k \cite{Imagenet_1k}, where SUN-324 is a subset of SUN-397 \cite{SUN397} as we remove 73 leaf classes with multiple parents in the label hierarchy. Similar to CuPL's settings on ImageNet, we remove \textit{`sunglass'} and \textit{`projectile'} in ImageNet-1k due to their noisy annotations confusing with \textit{`sunglasses'} and \textit{`missile'}, respectively. The datasets' statistics are summarized in Table~\ref{tab:datasets}. 
More details of the datasets and hierarchies are provided in the Appendix~\ref{sect:dataset_prep}.
We adopt three evaluation metrics from previous works \cite{BetterMistakes_hie_loss, BetterMistakes_CRM, HAFeature, HAFrame}: (1) Top1 accuracy (Top1), (2) average mistake severity (Severity), which is the sum of the hierarchical distances between incorrect predictions and their ground truth labels averaged over all incorrect predictions, and (3) HD@1, which is the hierarchical distance between top1 predictions and their ground truth labels averaged over all predictions. The hierarchical distance between two labels is defined as the height of their lowest common ancestor in the hierarchy. 

Our experiments focus on related methods that are training-free. We adopt CLIP's ensemble approach (dubbed CLIP in the remaining sections) as the baseline using the original image prompts provided by their GitHub Repository \cite{CLIP_GitHub} for each dataset. The test-time CRM predictions are based on CLIP's predictions. 
Following CLIP's official implementation \cite{CLIP_GitHub}, we scale its logit output by 100 so that CRM works appropriately. 
We also compare our approach with related CLIP-based zero-shot classification methods CuPL \cite{CuPL}, VCD \cite{VCD_LLM}, and HIE \cite{HieGroupComp} introduced in Sect.~\ref{sec:related_works}. 
The image prompts of all methods are produced with ChatGPT (`gpt-3.5-turbo-0125') \cite{chatGPT} following their respective query configurations.
All methods use the ViT-L/14@336px \cite{CLIP} image encoder in our experiments.
Since HIE is learning a label hierarchy via k-means clustering that synergizes with our hierarchy-aware approach, we also implement an alternative version where the label hierarchy is given to HIE. We dub the original HIE and our alternative version with HieC and HieT, respectively. 
For the score offset $\lambda$ in HIE, we search from 0.1 to 0.9 with step size 0.1 on the validation sets and select $\lambda$, which results in better HD@1 and Top1 accuracy (HD@1 has higher priority than Top1). 
Compared to the HIE variants, our approach directly compares the query class with the related classes given by the hierarchy instead of relying on LLM to generate a group summary for comparison. We also employ path-based prompts to stabilize the performance as introduced in Sect.~\ref{sect:language_prompts}. Lastly, we use the ensemble of image prompts' text embeddings for inference instead of score fusion. 
We summarize all methods employed in our experiments in Table~\ref{tab:method_summary}. More details on the implementations are provided in our code and Appendix~\ref{sect:methods_comp}-\ref{sect:Hie_config}.

\begin{figure}[t]
    \centering
    \includegraphics[height=2.2in]{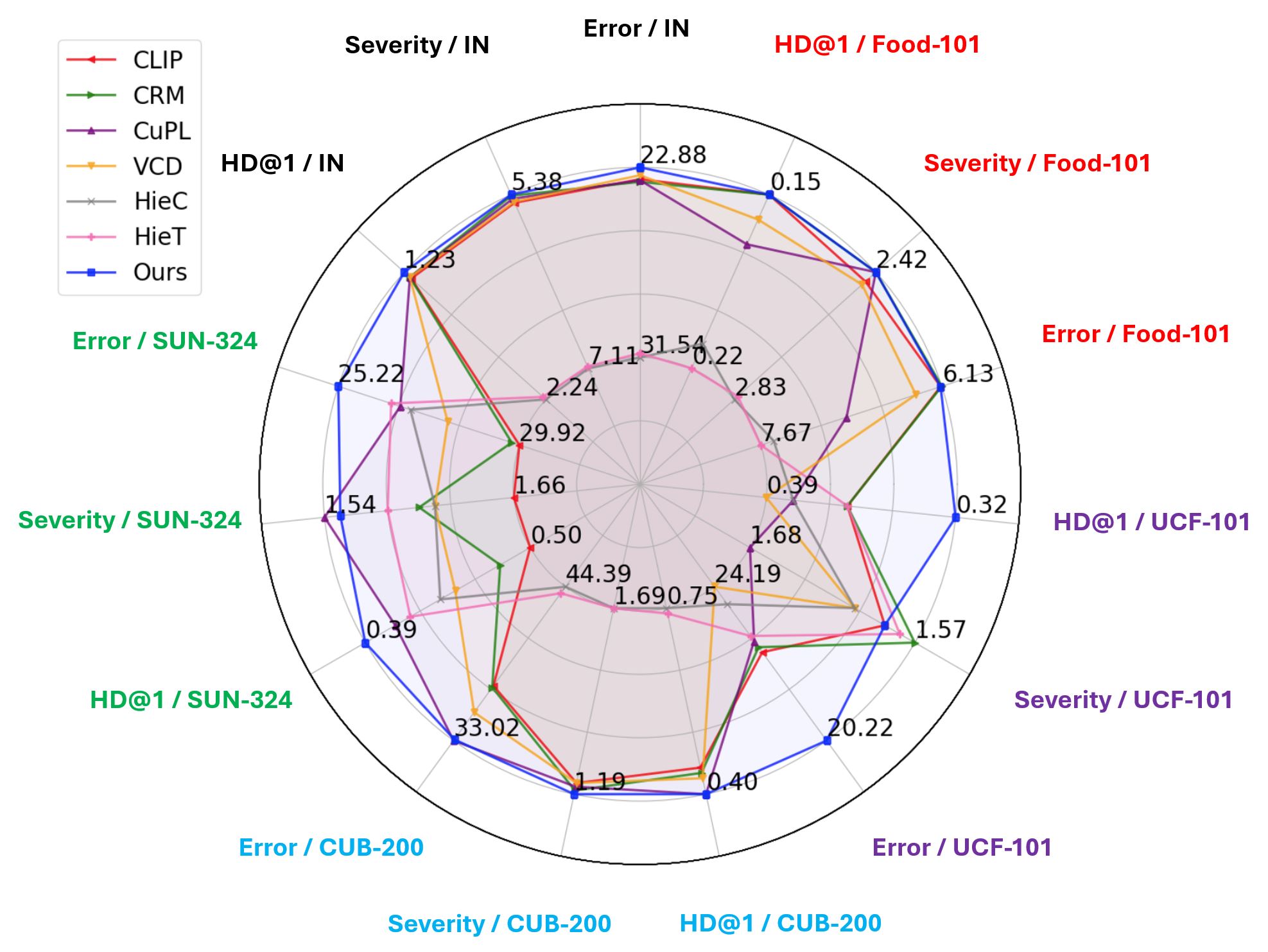}
    \caption{Holistic Comparison of all methods. We project the evaluation results of each metric/dataset pair to a specific polar axis in the radar chart. We convert Top1 accuracy to error rate so that all metrics are better when smaller (away from the origin). Larger polygons indicate better overall performances. `IN' represents ImageNet.}
    \label{fig:radar_main}
\end{figure}

\subsection{Results}

\begin{table*}[t]
    \centering
    \scriptsize
    \begin{tabular}{l|l|l|l}
    \hline
         Method & Language Prompts & Image Prompts & Inference\\
    \hline
         CLIP \cite{CLIP}   & NA              & Hand-crafted, shared image prompts & Ensemble over embedding space \\

         CRM \cite{BetterMistakes_CRM}   & NA              & NA                                 & Post-process CLIP predictions \\
         
         CuPL \cite{CuPL}  & Dataset-specific  & One image prompt per language prompt query + resampling & \makecell[tl]{Ensemble over embedding space}\\

         VCD \cite{VCD_LLM}   & Dataset-agnostic  & Multiple image prompts per language prompt query & \makecell[tl]{Ensemble over log-probability space}\\

         HIE (HieC \& HieT) \cite{HieGroupComp}    & Dataset-agnostic  & Multiple image prompts per language prompt query & \makecell[tl]{Multi-level score fusion}\\

         HAPrompts (ours) & Dataset-agnostic  & One image prompt per language prompt query & Ensemble over embedding space\\ 
         
    \hline
    \end{tabular}
    \caption{Comparison of all CLIP-based zero-shot classification methods.}
    \label{tab:method_summary}
\end{table*}

\begin{table}[h]
    \centering
    \scriptsize
    \begin{tabular}{c|l|c|c|c}
    \hline
    Dataset & Method & Top1$\uparrow$ & Severity$\downarrow$ & HD@1$\downarrow$ \\
         \hline 
         \multirow{7}{*}{\makecell{Food-101\\(height 4)}} 
         & CLIP & 93.86\% & 2.45 & \HB{0.15} \\
         & CRM & \HB{93.87\%} & \HB{2.42} & \HB{0.15} \\
         & CuPL & 93.06\% & \HB{2.42} & 0.17 \\
         & VCD & 93.65\% & 2.46 & 0.16 \\
         & HieC-0.3 & 92.44\% & 2.83 & 0.21 \\
         & HieT-0.5 & 92.33\% & 2.82 & 0.22 \\
         & Ours & 93.86\% & \HB{2.42} & \HB{0.15} \\

         \hline 
         \multirow{7}{*}{\makecell{UCF-101\\(height 3)}} 
         & CLIP & 77.50\% & 1.59 & 0.36 \\
         & CRM & 77.37\% & \HB{1.57} & 0.36 \\
         & CuPL & 77.21\% & 1.68 & 0.38 \\
         & VCD & 75.81\% & 1.61 & 0.39 \\
         & HieC-0.7 & 76.26\% & 1.61 & 0.38 \\
         & HieT-0.3 & 77.08\% & 1.58 & 0.36 \\
         & Ours & \HB{79.78\%} & 1.59 & \HB{0.32} \\

         \hline 
         \multirow{7}{*}{\makecell{CUB-200\\(height 3)}} 
         & CLIP & 62.93\% & 1.22 & 0.45 \\
         & CRM & 63.05\% & 1.20 & 0.44 \\
         & CuPL & \HB{66.98\%} & 1.21 & \HB{0.40} \\
         & VCD & 64.89\% & 1.22 & 0.43 \\
         & HieC-0.7 & 55.61\% & 1.69 & 0.75 \\
         & HieT-0.4 & 56.09\% & 1.69 & 0.74 \\
         & Ours & 66.90\% & \HB{1.19} & \HB{0.40} \\

         \hline 
         \multirow{7}{*}{\makecell{SUN-324\\(height 4)}} 
         & CLIP & 70.08\% & 1.66 & 0.50 \\
         & CRM & 70.28\% & 1.60 & 0.48 \\
         & CuPL & 73.17\% & \HB{1.54} & 0.41 \\
         & VCD & 71.94\% & 1.61 & 0.45 \\
         & HieC-0.6 & 72.91\% & 1.61 & 0.44 \\
         & HieT-0.3 & 73.41\% & 1.58 & 0.42 \\
         & Ours & \HB{74.78\%} & 1.55 & \HB{0.39} \\

         \hline 
         \multirow{7}{*}{\makecell{ImageNet\\(height 13)}} 
         & CLIP & 76.58\% & 5.46 & 1.28 \\
         & CRM & 76.46\% & 5.39 & 1.27 \\
         & CuPL & 76.53\% & 5.42 & 1.27 \\
         & VCD & 76.76\% & 5.45 & 1.27 \\
         & HieC-0.6 & 68.46\% & 7.11 & 2.24 \\
         & HieT-0.3 & 68.65\% & 7.09 & 2.22 \\
         & Ours & \HB{77.12\%} & \HB{5.38} & \HB{1.23} \\

         \hline 
         \multirow{7}{*}{\makecell{Average}} 
         & CLIP & 76.19\% & 2.48 & 0.55 \\
         & CRM & 76.21\% & 2.44 & 0.54 \\
         & CuPL & 77.51\% & 2.45 & 0.53 \\
         & VCD & 76.64\% & 2.47 & 0.54 \\
         & HieC & 73.14\% & 2.97 & 0.80 \\
         & HieT & 73.51\% & 2.95 & 0.79 \\
         & Ours & \HL{78.49\%} & \HL{2.43} & \HL{0.50} \\

    \hline
    \end{tabular}
    \caption{Evaluation results of all methods on the test sets. The best results among all methods for a dataset are highlighted in \colorbox{blue!15}{light purple}. The numbers after HieC and HieT methods are the $\lambda$ selected with validation sets.} 
    \label{tab:main_comp}
\end{table}

The holistic comparison of our approach with six methods on five datasets is shown as a radar chart in Fig.~\ref{fig:radar_main}, with the detailed results in Table~\ref{tab:main_comp}. Our approach reaches the best HD@1 on all five datasets. We also achieve the best average mistake severity and Top1 accuracy on 3/5 datasets. For Top1 accuracy on Food-101 and CUB-200, our approach is off by less than 0.1\% compared with the best results while maintaining the best mistake severity and HD@1. For average mistake severity on UCF-101 and SUN-324, our performance drops 0.02 and 0.01 compared to the best results while maintaining the best HD@1 and improving the respective Top1 accuracy by 2\% and 1\% compared with the second-best approach. As shown in the radar chart in Fig.~\ref{fig:radar_main}, our approach (\textcolor{blue}{blue polygon}) achieves a better balance between mistake severity and classification accuracy, outperforming other methods in the holistic comparison. This conforms with the average evaluation results across five datasets in the bottom rows of Table~\ref{tab:main_comp}. Compared with our approach, the other two hierarchy-aware methods, CRM and HieT, are less robust against datasets of varying scales with hierarchies of different heights. Notably, the performance of both HieC and HieT are unstable as their accuracies drop significantly on CUB-200 and ImageNet compared with the other methods. In Appendix~\ref{sect:histogram_full}, we provide histograms of mistake severities for all methods on the test set of each dataset. Our method produces fewer high-severity mistakes than other approaches, except for SUN-324.

\subsection{Ablation Study}

Our proposed approach includes three different groups of language prompts: (1) leaf-peer comparative prompts ($\mathcal{P}^{LP}$), (2) ancestor-peer comparative prompts ($\mathcal{P}^{AP}$), and (3) path-based generic prompts ($\mathcal{P}^{G}$). Our ablation study includes all seven combinations of the above three groups of language prompts. We evaluate their performance on the validation sets of the five datasets employed in our study. The holistic comparison of our ablation study on five datasets is presented in Fig.~\ref{fig:radar_ablation}, with detailed results in Appendix~\ref{sect:ablation_full}. The leaf-peer comparative prompts ($\mathcal{P}^{LP}$, \textcolor{red}{red polygon}) have better severity on ImageNet and UCF-101 at the cost of significantly lower HD@1 and Top1 accuracy, i.e., it makes many more low-severity mistakes to drive down its average mistake severity. The comparative prompts ($\mathcal{P}^{LP}\cup \mathcal{P}^{AP}$, \textcolor{orange}{orange polygon}) achieve slightly better performance than the proposed approach in 3/15 of the comparisons. The path-based generic prompts ($\mathcal{P}^{G}$, \textcolor{purple}{purple polygon}) are completely outperformed by our approach ($\mathcal{P}^{LP}\cup \mathcal{P}^{AP}\cup \mathcal{P}^{G}$, \textcolor{blue}{blue polygon}). In summary, our approach has the best overall results.

\begin{figure}[t]
    \centering
    \includegraphics[height=2in]{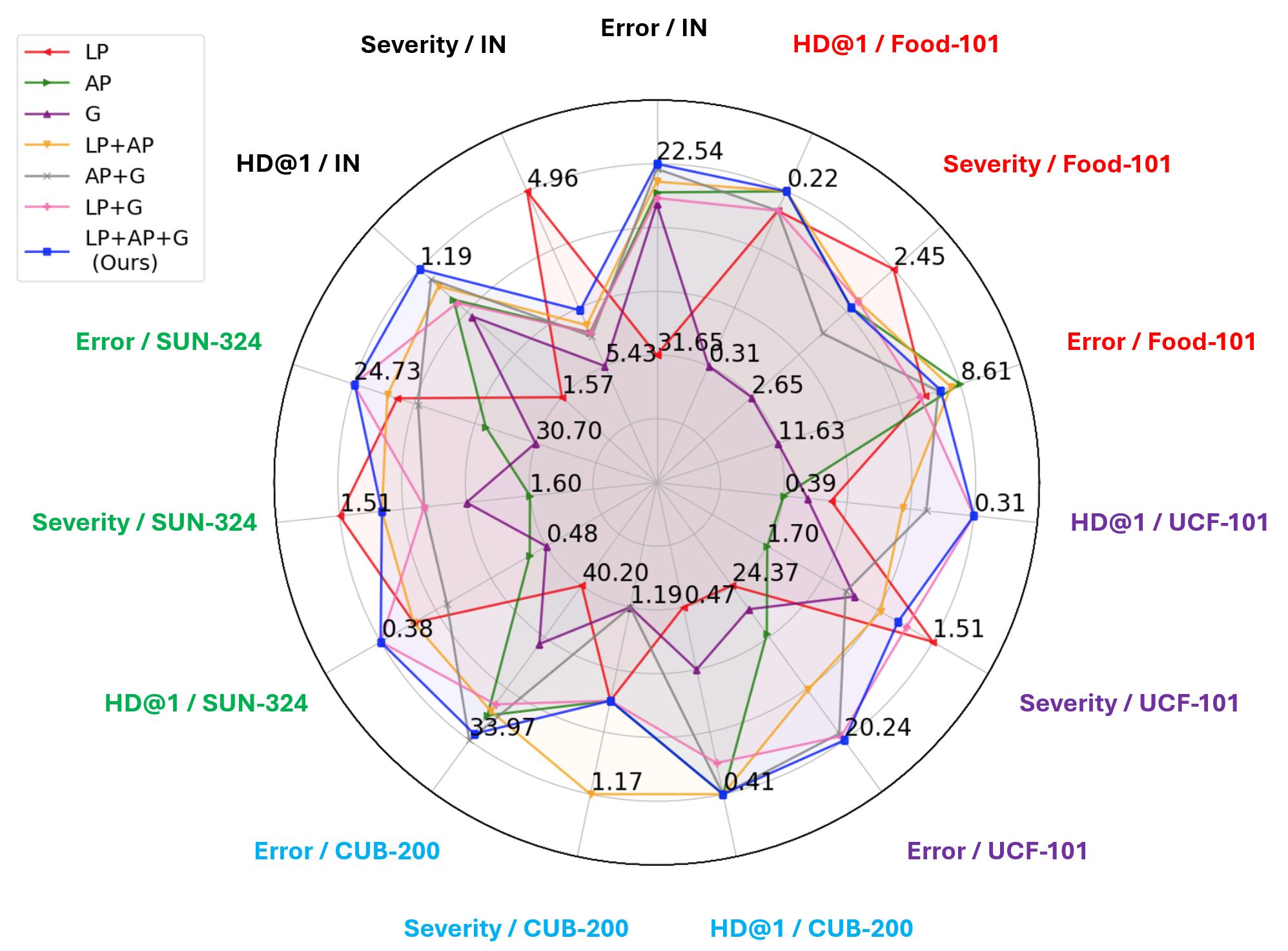}
    \caption{Holistic comparison of ablation study results.}
    \label{fig:radar_ablation}
\end{figure}

\subsection{Ensemble of Image Prompts}
\begin{table}[b]
    \centering
    \scriptsize
    \begin{tabular}{c|l|c|c|c}
    \hline
    Dataset & Ensemble & Top1$\uparrow$ & Severity$\downarrow$ & HD@1$\downarrow$ \\
         \hline 
         \multirow{2}{*}{\makecell{Food-101}}
         &  log-prob  & \textcolor{red}{91.30\%} & 2.51 & 0.22 \\
         &  embedding & 91.07\% & 2.51 & 0.22 \\
         
         \hline
         \multirow{2}{*}{\makecell{UCF-101}}
         &  log-prob  & 76.79\% & \textcolor{red}{1.49} & 0.35 \\
         &  embedding & 79.76\% & 1.55 & 0.31 \\
         
         \hline
         \multirow{2}{*}{\makecell{CUB-200}}
         &  log-prob  & \textcolor{red}{66.05\%} & 1.18 & \textcolor{red}{0.40} \\
         &  embedding & 65.77\% & 1.18 & 0.41 \\

         \hline
         \multirow{2}{*}{\makecell{SUN-324}}
         &  log-prob  & 74.07\% & \textcolor{red}{1.51} & 0.39 \\
         &  embedding & 75.23\% & 1.53 & 0.38 \\

         \hline
         \multirow{2}{*}{\makecell{ImageNet}}
         &  log-prob  & 77.19\% & 5.30 & 1.21 \\
         &  embedding & 77.46\% & 5.28 & 1.19 \\
         \hline
         
    \end{tabular}
    \caption{Comparison results of ensemble methods evaluated on the validation set of all datasets using our language prompts.}
    \label{tab:ensemble}
\end{table}
In our approach, we ensemble the text embeddings derived from image prompts of the same class to form its overall class text embedding, the same as CLIP and CuPL. We compare our ensemble over the embedding space with VCD's ensemble over log-probability (logit) space, i.e., the text embedding of each image prompt serves as a sub-classifier. The logits of sub-classifiers belonging to the same class are averaged to produce the logit of the respective class. The results on validation sets are shown in Table~\ref{tab:ensemble}. We highlighted results where the log-probability ensemble outperforms our approach in \textcolor{red}{red}. Our approach suffers minor accuracy drops ($\le0.3\%$) on Food-101 and CUB-200. In UCF-101 and SUN-324, ensemble over log-probability space has better mistake severity at a cost of $1\%\sim$3\% accuracy drop. The ensemble over the embedding space leads to overall better performance.

\subsection{Transferability of Language Prompts}
\begin{table}[t]
    \centering
    \scriptsize
    \begin{tabular}{l|l|l|l}
    \hline
         LLM & Top1$\uparrow$ & Severity$\downarrow$ & HD@1$\downarrow$ \\
         \hline
          CLIP & 76.58\% & 5.46 & 1.28 \\
          CRM & 76.46\% & 5.39 & 1.27 \\
          \hline
          ChatGPT & 77.12\% & 5.38 & 1.23 \\
          Claude & 76.45\% (\textcolor{red}{-0.67\%})  & 5.36 (\textcolor{blue!100}{-0.02}) & 1.26 (\textcolor{red}{+0.03})\\
          Gemini & 76.95\% (\textcolor{red}{-0.17\%}) & 5.38 & 1.24 (\textcolor{red}{+0.01})\\
    \hline
    \end{tabular}
    \caption{Transferability of the proposed language prompts with different LLMs evaluated on ImageNet.}
    \label{tab:language_prompt_transfer}
\end{table}
We test the transferability of our proposed language prompts to different foundational LLMs. CLIP and CRM serve as references to the expected zero-shot performance on ImageNet. We feed the same language prompts introduced in Sect.~\ref{sect:language_prompts} to Claude-3.5-sonnet \cite{Claude} and Gemini-1.5-flash \cite{Gemini} to generate the associated image prompts for subsequent evaluation of their zero-shot performance. The results are shown in Table~\ref{tab:language_prompt_transfer}. The Top1 accuracy only drops 0.67\% and 0.17\% when switching from ChatGPT to Claude and Gemini, respectively. Meanwhile, the average mistake severity remains at 5.38 or below and the HD@1 remains lower than 1.27 (HD@1 of CRM). 
This demonstrates the robustness of our proposed language prompts against changes of the underlying language model.

\section{Limitation}
Our method relies on the LLM to generate image prompts based on our language prompts. The resulting image prompts are directly used to produce the text embeddings for zero-shot classification. This does not guarantee that the generated image prompts are free from stereotypical descriptions about the class being queried, which may lead to a negative social impact. 
Though one could manually edit the resulting improper image prompts, it is labor intensive. Future work may employ an additional moderation model to examine and adjust the image prompts instead.

\section{Conclusion}
We propose to incorporate prior knowledge of a label hierarchy into language prompts through structured comparison between classes in the hierarchy and embedded hints of ancestor classes. The resulting image prompts from querying the LLM enhance the class text embeddings to better reflect hierarchical semantic relationships among leaf classes.
A holistic comparison with six methods on five datasets demonstrates that our approach outperforms others in the tradeoff between classification accuracy and making better mistakes.

{
    \small
    \bibliographystyle{ieeenat_fullname}
    \bibliography{main}
}

\clearpage
\setcounter{page}{1}
\maketitlesupplementary

\section{Appendix}
\label{sec:appendix}
%
\begin{table*}[t]
    \centering
    \scriptsize
    \begin{tabular}{l|l|l|l}
    \hline
         Method & Language Prompts & Image Prompts & Inference\\
    \hline
         CLIP \cite{CLIP}   & NA              & Hand-crafted, shared image prompts & Ensemble over embedding space \\

         CRM \cite{BetterMistakes_CRM}   & NA              & NA                                 & Post-process CLIP predictions \\
         
         CuPL \cite{CuPL}  & Dataset-specific  & One image prompt per language prompt query + resampling & \makecell[tl]{Ensemble over embedding space}\\

         VCD \cite{VCD_LLM}   & Dataset-agnostic  & Multiple image prompts per language prompt query & \makecell[tl]{Ensemble over log-probability space}\\

         HIE (HieC \& HieT) \cite{HieGroupComp}    & Dataset-agnostic  & Multiple image prompts per language prompt query & \makecell[tl]{Multi-level score fusion}\\

         HAPrompts (ours) & Dataset-agnostic  & One image prompt per language prompt query & Ensemble over embedding space\\ 
         
    \hline
    \end{tabular}
    \caption{Comparison of all CLIP-based zero-shot classification methods.}
    \label{tab:method_summary_appendix}
\end{table*}

We provide more details of dataset preparation and the associated hierarchy in Sect.~\ref{sect:dataset_prep}, a comparison between different methods in Sect.~\ref{sect:methods_comp}, and the configuration employed to query LLM for all methods in our experiments in Sect.~\ref{sect:LLM_config}. We also detail our alternative implementation of HIE \cite{HieGroupComp} (HieT) conditioned on the label hierarchy and the hyper-parameter settings for both HieC and HieT in Sect.~\ref{sect:Hie_config}. The detailed results of our ablation study are provided in Sect.~\ref{sect:ablation_full}. The histograms of mistake severities for all methods on all five datasets are shown in Sect.~\ref{sect:histogram_full}. Lastly, we show some examples our correct predictions, the associated test image, and ground truth image prompt examples in Sect.~\ref{sect:visual_example}.

\subsection{Datasets and Label Hierarchies} \label{sect:dataset_prep}
\begin{figure*}
    \centering
    \includegraphics[height=5.0in]{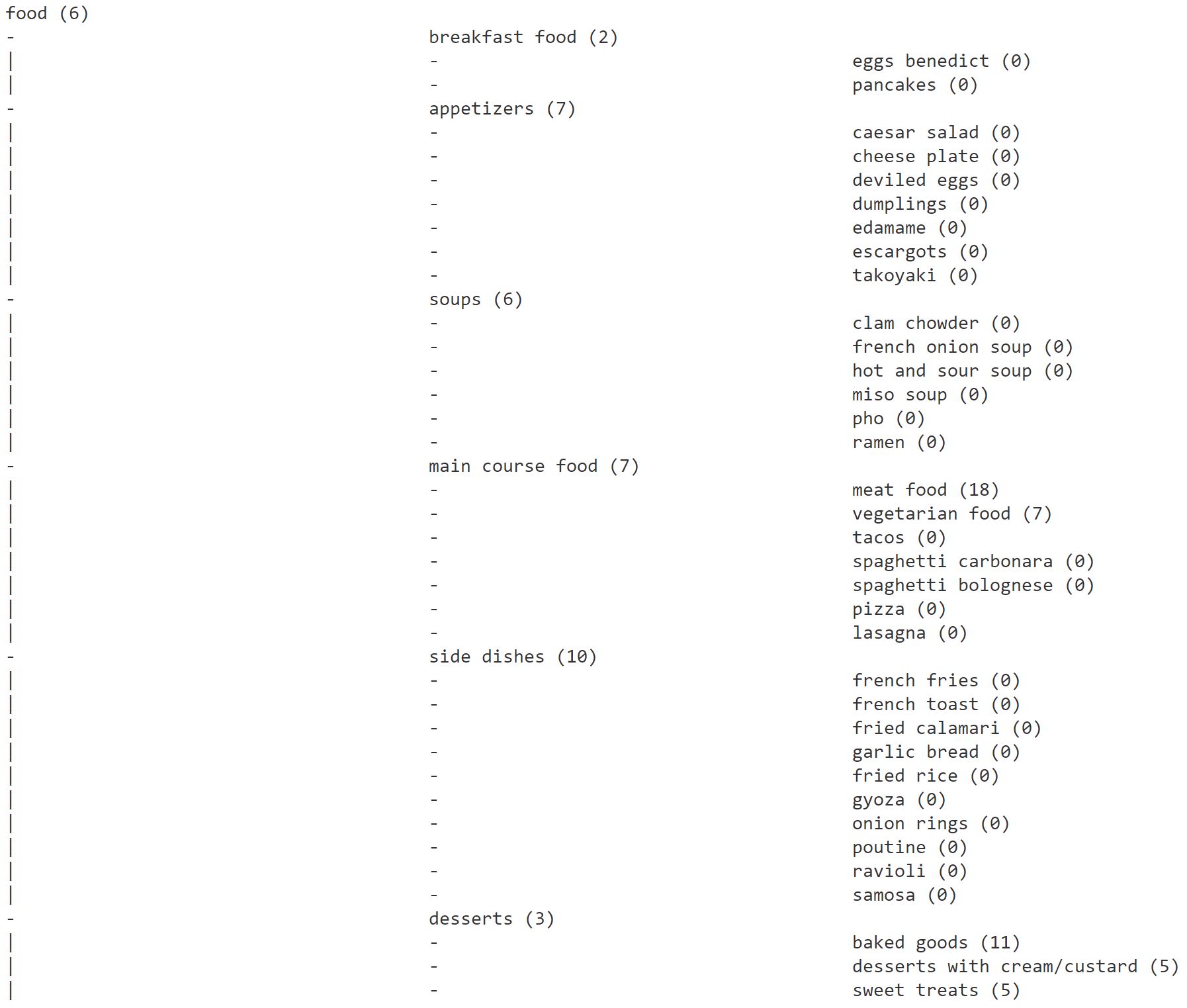}
    \caption{The first three levels of the label hierarchy for Food-101. The number inside the parenthesis after each node label indicates the number of direct children the associated node has in the hierarchy.}
    \label{fig:food_101_tree}
\end{figure*}

\begin{figure*}
    \centering
    \includegraphics[height=5.8in]{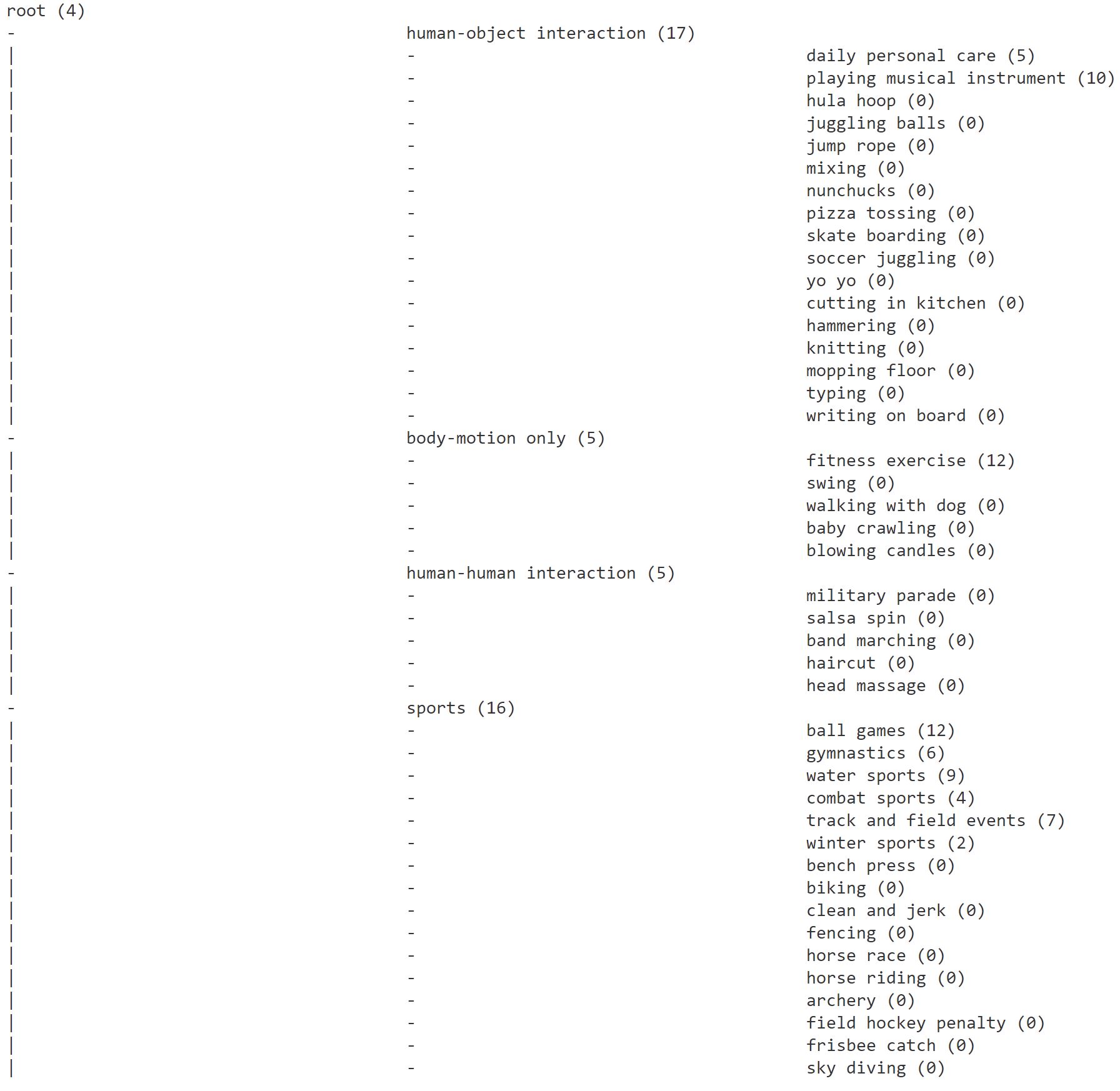}
    \caption{The first three levels of the label hierarchy for UCF-101. The number inside the parenthesis after each node label indicates the number of direct children the associated node has in the hierarchy.}
    \label{fig:ucf_101_tree}
\end{figure*}

\begin{figure*}
    \centering
    \includegraphics[height=6.6in]{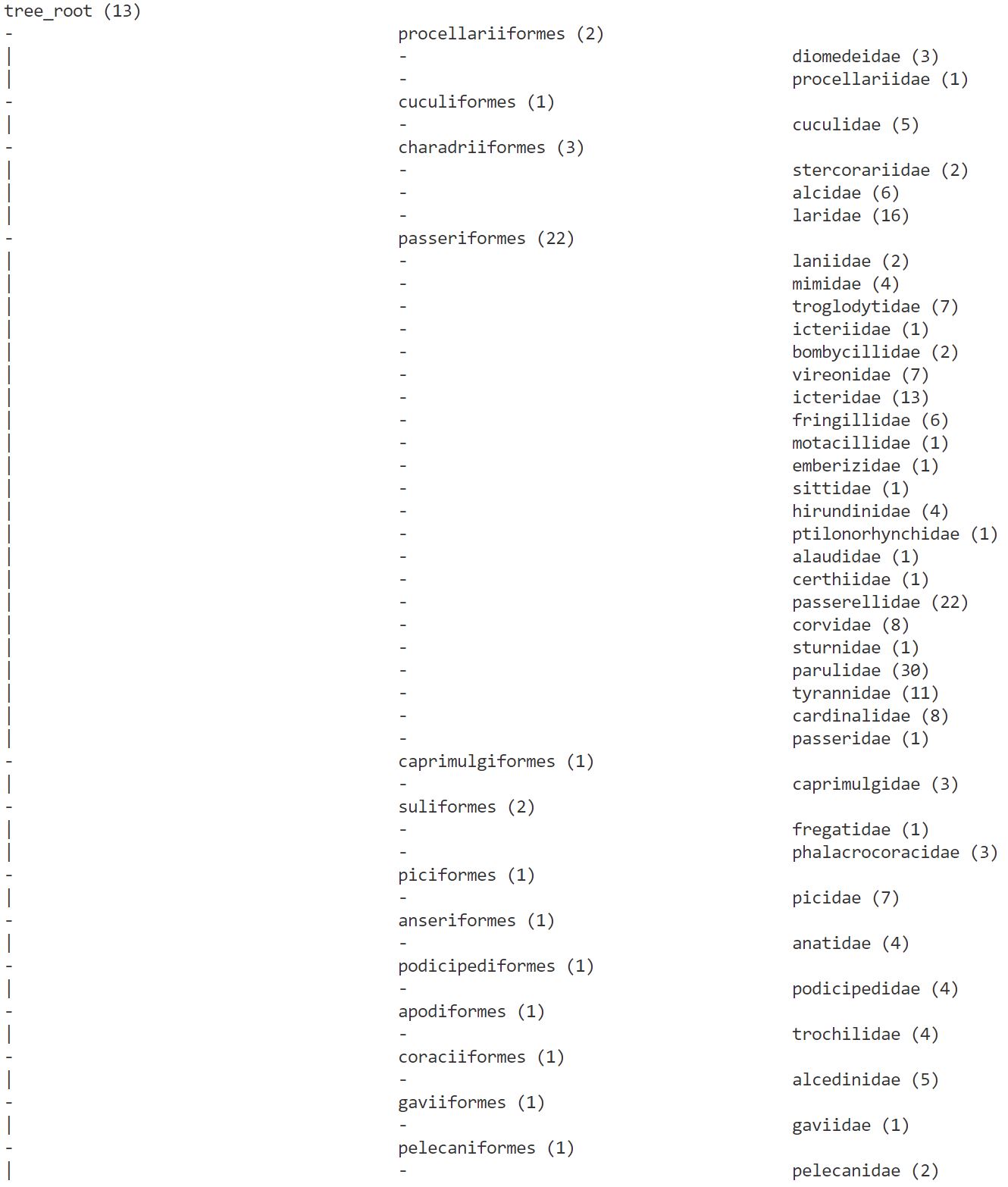}
    \caption{The first three levels of the label hierarchy for CUB-200. The number inside the parenthesis after each node label indicates the number of direct children the associated node has in the hierarchy.}
    \label{fig:cub_200_tree}
\end{figure*}

\begin{figure*}
    \centering
    \includegraphics[height=1.2in]{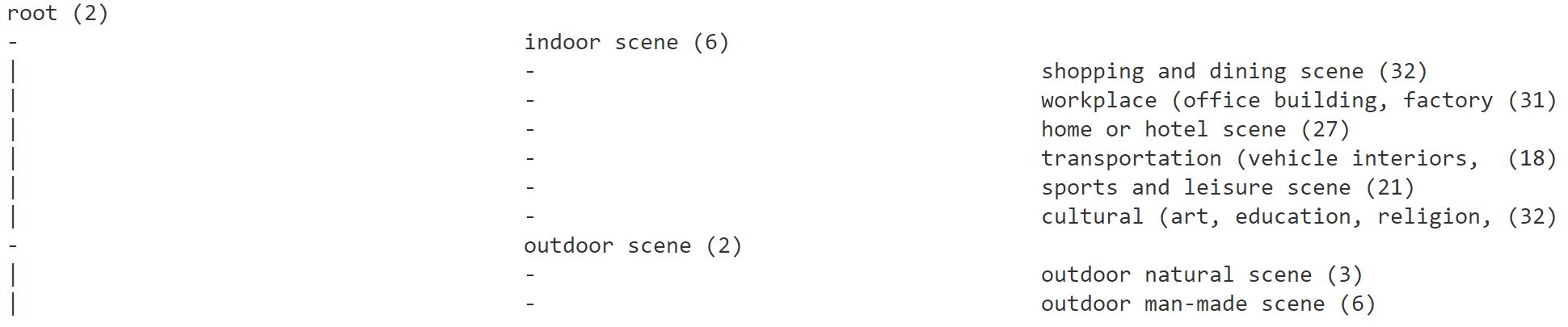}
    \caption{The first three levels of the label hierarchy for SUN-324. The number inside the parenthesis after each node label indicates the number of direct children the associated node has in the hierarchy. Some node labels are truncated in this visualization due to space limit.}
    \label{fig:sun_324_tree}
\end{figure*}

\begin{figure*}
    \centering
    \includegraphics[height=4.6in]{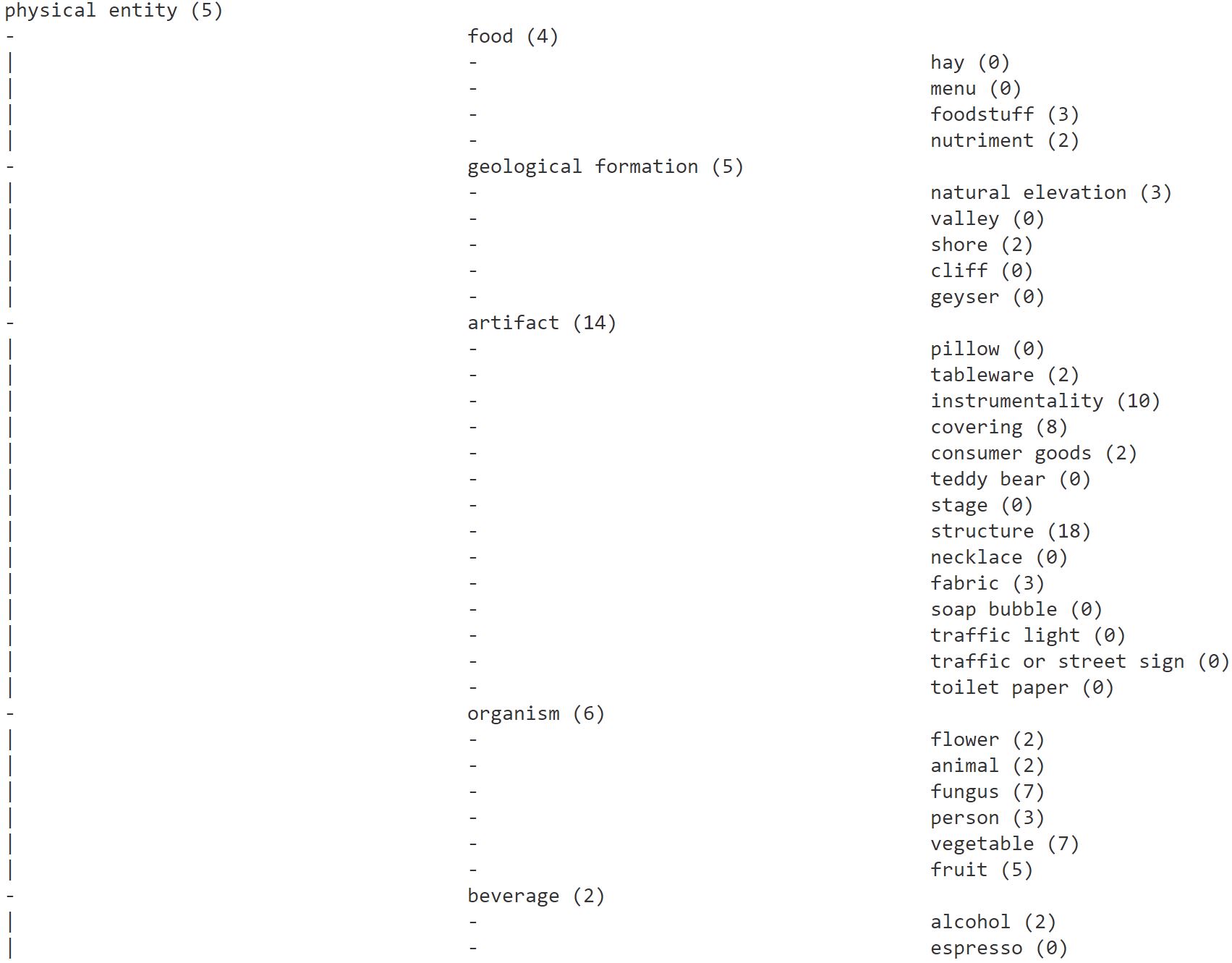}
    \caption{The first three levels of the label hierarchy for ImageNet. The number inside the parenthesis after each node label indicates the number of direct children the associated node has in the hierarchy.}
    \label{fig:imagenet_tree}
\end{figure*}

We cover the dataset preparation for our comparative experiments, which includes constructing our validation and test sets, modifying certain class names, or removing confusing classes with incorrect annotations for the respective datasets. 
We provide code to reconstruct the partitions of all datasets adopted in our experiments and the associated files to represent the label hierarchies. Due to the sheer size of the associated label hierarchies, we only provide partial views (first three levels) of the associated label hierarchies in Fig.~\ref{fig:food_101_tree} to Fig.~\ref{fig:imagenet_tree}, the full views of these hierarchies are in our Github repo at \href{https://github.com/ltong1130ztr/HAPrompts/tree/master/tree_viz}{tree visualization link}.

\noindent\textbf{Food-101.}
The original Food-101 \cite{food101} dataset contains 750 training examples per class and 250 testing examples per class. We keep the original test set and randomly sample 250 examples per class from the training set as validation examples in our experiments.
We acquire Food-101's label hierarchy by querying ChatGPT to organize the associated classes into a tree structure first, and then manually adjusting them. 

\noindent\textbf{UCF-101.}
The original UCF-101 \cite{ucf101} video dataset contains three train-test splits of the same examples. We employ its first training split as our validation set and its first test split as our test set. For each video clip, we extract its middle frame as its image example for the clip to construct our UCF-101 image dataset. We acquire UCF-101's label hierarchy by asking ChatGPT to refine the original simple hierarchy provided by UCF-101.

\noindent\textbf{CUB-200.}
We fetch 20 examples per class from the original training set of CUB-200 \cite{cub200} as our validation set and adopt the original test set as our test set in the experiment. 
We adopt the label hierarchy from \cite{flamingo_cub200_tree}.

\noindent\textbf{SUN-324.}
The SUN-324 dataset is a subset of the SUN-397 \cite{SUN397} dataset. We removed 73 classes from SUN-397 since these classes have ambiguous hierarchical labels, i.e., the same leaf class has multiple parents. The original SUN-397 dataset does not provide a train-val-test split. Therefore, we sample 25 validation examples per class and 50 test examples per class from the remaining 324 classes in SUN-397 to form our SUN-324 dataset. We use the original label hierarchy provided by SUN-397 without the 73 classes that have multiple parents in the hierarchy. 

\noindent\textbf{ImageNet-1k.}
We remove two classes in the original ImageNet-1k \cite{Imagenet_1k} classes as their annotations are either incorrect or confusing with other classes: (1) \textit{`sunglass'} (n04355933) with the definition: `a convex lens that focuses the rays of the sun; used to start a fire'. But its examples are \textit{`sunglasses'} that are `darkened or polarized to protect the eyes from the glare of the sun'; (2) \textit{`projectile'} (n04008634) with the definition: `a weapon that is forcibly thrown or projected at a target but is not self-propelled' which confuses with \textit{`missile'}, both classes include self-propelled missile examples. This practice is similar to the preprocessing in CuPL \cite{CuPL}, which merges the above two pairs of classes. In addition, we employ the class names used in CuPL with two changes adopted from VCD: (1) \textit{`tights'} $\rightarrow$ \textit{`maillot'} (2) \textit{`newt'} $\rightarrow$ \textit{`eft'} and make one change of our own: \textit{`bubble'} $\rightarrow$ \textit{`soap bubble'}. The updated class names are applied to all related methods in our experiments to query LLM for the respective image prompts. For the remaining 998 classes in ImageNet-1k, we split the original validation set with 50 examples per class into our validation and test sets with a 1:1 ratio. We derive the label hierarchy from the associated classes' WordNet \cite{WordNet} synsets. We follow the approach employed in \cite{BetterMistakes_hie_loss} to build our initial hierarchy from WordNet. Then, we manually trimmed some highly abstract classes in the hierarchy, such as \textit{`matter'}, \textit{`object'}, \textit{`abstraction'}, and \textit{`whole'}.

\subsection{Methods Comparison} \label{sect:methods_comp}
In Table~\ref{tab:method_summary_appendix}, we summarize different methods, including the proposed HAPrompts approach. As introduced in Sect.~\ref{sect:language_prompts}, our hierarchy-aware language prompts use the same templates for all datasets, we merely inject hierarchical information derived from the associated label hierarchy. The HIE and VCD approaches also use the same language prompt for all datasets, whereas CuPL has hand-crafted language prompts for each dataset. Both CuPL and our approach directly use the entire response of the LLM to one language prompt as the associated image prompt, unlike VCD and HIE approaches, where they decompose the LLM response for one language prompt into multiple image prompts. The LLM response decomposition is achieved by inducing the language model to break its response into numerous bullet points. Both VCD and HIE provide a question-answer example in their language prompts (in-context learning) and insert the \textcolor{blue}{``-''} symbol at the end of the prompt. The bullet points are then extracted through a hard-coded text-processing step to form coherent sentences as the final image prompts.

During inference, our HAPrompts and CuPL ensemble the text embedding of image prompts for each leaf-level class to produce the associated class text embedding (classifier weight) similar to CLIP for the subsequent inner product with the test image embedding and, therefore, acquire the logits for classification. On the other hand, VCD employs ensemble over the log-probability (logit) space, i.e., treating each image prompt for the same class as a sub-classifier to compute the inner product between sub-classifier weight and test image embedding and then averaging the resulting logits to acquire the final logit for the class. The HIE approaches employ a more sophisticated multi-level score fusion approach based on the associated hierarchy to compute the logit for each leaf class instead. 

\subsection{Configuration for LLM Query} \label{sect:LLM_config}
All methods that use language prompts to query LLM for image prompts use the same underlying ChatGPT \cite{chatGPT} language model (`gpt-3.5-turbo-0125'). We follow the original query configuration for each method and provide associated details and our approach's configuration in Table.~\ref{tab:query_config}. 

\begin{table}[]
    \centering
    \scriptsize
    \begin{tabular}{l|l|l|l}
    \hline
         Language Prompts & max\_token & stop\_token & temperature \\
         \hline
         CuPL  & 50 & `.' & 0.99 \\
         VCD   & 100 & None & 0.00 \\
         HieC  & None & None & 1.00 \\
         HieT  & None & None & 1.00 \\
         $\mathcal{P}^{LP}$ (Ours) & 150 & `.' & 1.00 \\
         $\mathcal{P}^{LP}$ (Ours) & 150 & `.' & 1.00 \\
         $\mathcal{P}^{G}$ (Ours) & 150 & None & 1.00 \\
    \hline
    \end{tabular}
    \caption{Configuration employed for each method to query LLM with language prompts.}
    \label{tab:query_config}
\end{table}

\subsection{HIE Approach Implementation Details} \label{sect:Hie_config}
The original HIE \cite{HieGroupComp} approach (i.e., HieC in our experiments) is building a label hierarchy on the fly with iterative k-means clustering of intermediate image prompts for classes grouped into the same cluster. The associated language prompts compare classes within the same cluster to produce finer image prompts for subsequent clustering till the cluster size is reduced to a threshold $l$ or smaller. The authors suggest setting $l$ to 2 or 3 but do not provide the exact number employed in their experiments. This is also the case for the number of clusters $n$ in k-means clustering. Following their Github repository's code example, we set $l=3$ and $n=6$ for all datasets. 
\subsection{More Details of Our Ablation Study} \label{sect:ablation_full}
We exhaust all seven combinations of our proposed three types of language prompts: leaf-peer comparative prompts ($\mathcal{P}^{LP}$), ancestor-peer comparative prompts ($\mathcal{P}^{AP}$), and path-based generic prompts ($\mathcal{P}^G$). The evaluation results of each variant on the validation set of the five datasets employed in our experiments are shown in Table.~\ref{tab:ablation_comp}. The image prompts for $\mathcal{P}^{LP}$, $\mathcal{P}^{AP}$, and $\mathcal{P}^{LP}\cup \mathcal{P}^{LP}$ are generated by LLM separately. 
For $\mathcal{P}^{LP}$, there are corner cases where a leaf class is connected to the root directly. Hence, it has no leaf-level peers. We use their ancestor peers to generate the respective image prompts. 
Our approach ($\mathcal{P}^{LP}\cup \mathcal{P}^{AP}\cup \mathcal{P}^{G}$) reaches best Top1 accuracy and HD@1 averaged across five datasets, and its average mistake severity is only second to the leaf-peer comparative prompts ($\mathcal{P}^{LP}$), which is not stable and achieves better mistake severity while suffering significant accuracy drop in CUB-200 ($-5\%$) and ImageNet ($-4\%$).

\begin{table}[t]
    \centering
    \scriptsize
    \begin{tabular}{S|T|S|S|S}
    \hline
    Dataset & Variants & Top1$\uparrow$ & Severity$\downarrow$ & HD@1$\downarrow$ \\
         \hline 
         \multirow{7}{*}{\makecell{Food-101\\(height 4)}} 
         & $\mathcal{P}^{LP}$ & 90.81\% & 2.45 & 0.23 \\
         & $\mathcal{P}^{AP}$ & 91.39\% & 2.51 & 0.22 \\
         & $\mathcal{P}^{G}$ & 88.37\% & 2.65 & 0.31 \\
         & $\mathcal{P}^{LP}\cup \mathcal{P}^{AP}$ & 91.24\% & 2.50 & 0.22 \\
         & $\mathcal{P}^{AP}\cup \mathcal{P}^{G}$ & 91.03\% & 2.55 & 0.23 \\
         & $\mathcal{P}^{LP}\cup \mathcal{P}^{G}$ & 90.73\% & 2.50 & 0.23 \\
         & $\mathcal{P}^{LP}\cup \mathcal{P}^{AP}\cup \mathcal{P}^{G}$ & 91.07\%  & 2.51 & 0.22 \\

         \hline 
         \multirow{7}{*}{\makecell{UCF-101\\(height 3)}} 
         & $\mathcal{P}^{LP}$ & 75.63\% & 1.51 & 0.37 \\
         & $\mathcal{P}^{AP}$ & 76.92\% & 1.70 & 0.39 \\
         & $\mathcal{P}^{G}$ & 76.26\% & 1.60 & 0.38 \\
         & $\mathcal{P}^{LP}\cup \mathcal{P}^{AP}$ & 78.42\% & 1.57 & 0.34 \\
         & $\mathcal{P}^{AP}\cup \mathcal{P}^{G}$ & 79.58\% & 1.61 & 0.33 \\
         & $\mathcal{P}^{LP}\cup \mathcal{P}^{G}$ & 79.65\% & 1.54 & 0.31 \\
         & $\mathcal{P}^{LP}\cup \mathcal{P}^{AP}\cup \mathcal{P}^{G}$ & 79.76\% & 1.55 & 0.31 \\

         \hline 
         \multirow{7}{*}{\makecell{CUB-200\\(height 3)}} 
         & $\mathcal{P}^{LP}$ & 59.80\% & 1.18 & 0.47 \\
         & $\mathcal{P}^{AP}$ & 65.03\% & 1.18 & 0.41 \\
         & $\mathcal{P}^{G}$ & 62.15\% & 1.19 & 0.45 \\
         & $\mathcal{P}^{LP}\cup \mathcal{P}^{AP}$ & 64.85\% & 1.17 & 0.41 \\
         & $\mathcal{P}^{AP}\cup \mathcal{P}^{G}$ & 66.03\% & 1.19 & 0.41 \\
         & $\mathcal{P}^{LP}\cup \mathcal{P}^{G}$ & 64.58\% & 1.18 & 0.42 \\
         & $\mathcal{P}^{LP}\cup \mathcal{P}^{AP}\cup \mathcal{P}^{G}$ & 65.77\% & 1.18 & 0.41 \\

         \hline 
         \multirow{7}{*}{\makecell{SUN-324\\(height 4)}} 
         & $\mathcal{P}^{LP}$ & 73.83\% & 1.51 & 0.40 \\
         & $\mathcal{P}^{AP}$ & 70.93\% & 1.60 & 0.47 \\
         & $\mathcal{P}^{G}$ & 69.30\% & 1.57 & 0.48 \\
         & $\mathcal{P}^{LP}\cup \mathcal{P}^{AP}$ & 74.16\% & 1.53 & 0.40 \\
         & $\mathcal{P}^{AP}\cup \mathcal{P}^{G}$ & 73.15\% & 1.55 & 0.42 \\
         & $\mathcal{P}^{LP}\cup \mathcal{P}^{G}$ & 75.27\% & 1.55 & 0.38 \\
         & $\mathcal{P}^{LP}\cup \mathcal{P}^{AP}\cup \mathcal{P}^{G}$ & 75.23\% & 1.53 & 0.38 \\

         \hline 
         \multirow{7}{*}{\makecell{ImageNet\\(height 13)}} 
         & $\mathcal{P}^{LP}$ & 68.35\% & 4.96 & 1.57 \\
         & $\mathcal{P}^{AP}$ & 76.10\% & 5.34 & 1.28 \\
         & $\mathcal{P}^{G}$ & 75.54\% & 5.43 & 1.33 \\
         & $\mathcal{P}^{LP}\cup \mathcal{P}^{AP}$ & 76.61\% & 5.32 & 1.24 \\
         & $\mathcal{P}^{AP}\cup \mathcal{P}^{G}$ & 77.22\% & 5.35 & 1.22 \\
         & $\mathcal{P}^{LP}\cup \mathcal{P}^{G}$ & 75.82\% & 5.34 & 1.29 \\
         & $\mathcal{P}^{LP}\cup \mathcal{P}^{AP}\cup \mathcal{P}^{G}$ & 77.46\% & 5.28 & 1.19 \\

         \hline 
         \multirow{7}{*}{\makecell{Average}} 
         & $\mathcal{P}^{LP}$ & 73.68\% & \HB{2.32} & 0.61 \\
         & $\mathcal{P}^{AP}$ & 76.07\% & 2.47 & 0.55 \\
         & $\mathcal{P}^{G}$ & 74.32\% & 2.49 & 0.59 \\
         & $\mathcal{P}^{LP}\cup \mathcal{P}^{AP}$ & 77.06\% & 2.42 & 0.52 \\
         & $\mathcal{P}^{AP}\cup \mathcal{P}^{G}$ & 77.40\% & 2.45 & 0.52 \\
         & $\mathcal{P}^{LP}\cup \mathcal{P}^{G}$ & 77.21\% & 2.42 & 0.53 \\
         & $\mathcal{P}^{LP}\cup \mathcal{P}^{AP}\cup \mathcal{P}^{G}$ & \HB{77.86\%} & 2.41 & \HB{0.50} \\
         
    \hline
    \end{tabular}
    \caption{Evaluation of all variants in ablation study on the validation set. The best average results in the bottom rows among all variants are highlighted in \colorbox{blue!15}{light purple}. } 
    \label{tab:ablation_comp}
\end{table}

\begin{figure*}
\centering
\setlength{\tabcolsep}{8pt}
    \begin{tabular}{c c}
    \small
    Food-101 & UCF-101 \\
    \includegraphics[height=1.2in]{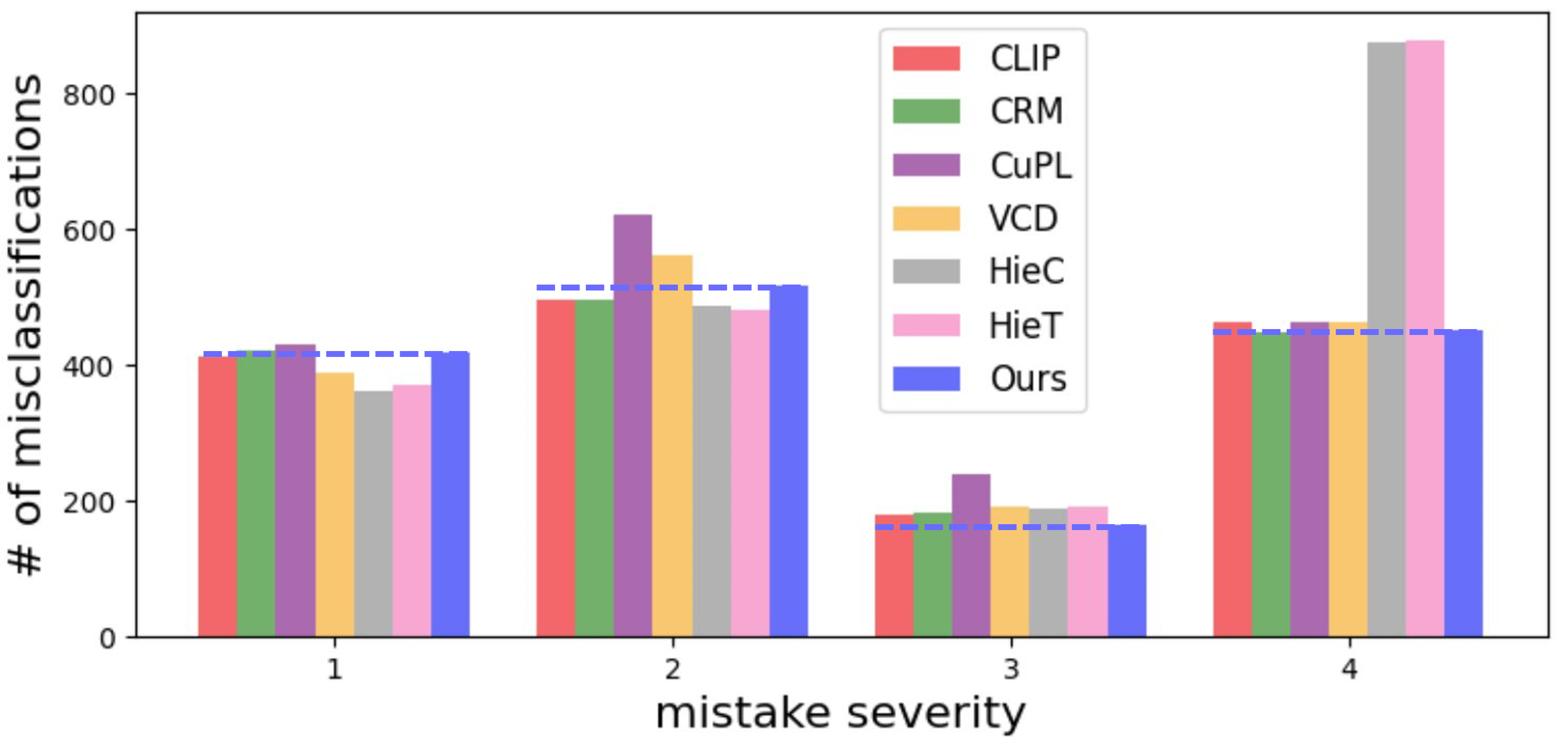} &  \includegraphics[height=1.2in]{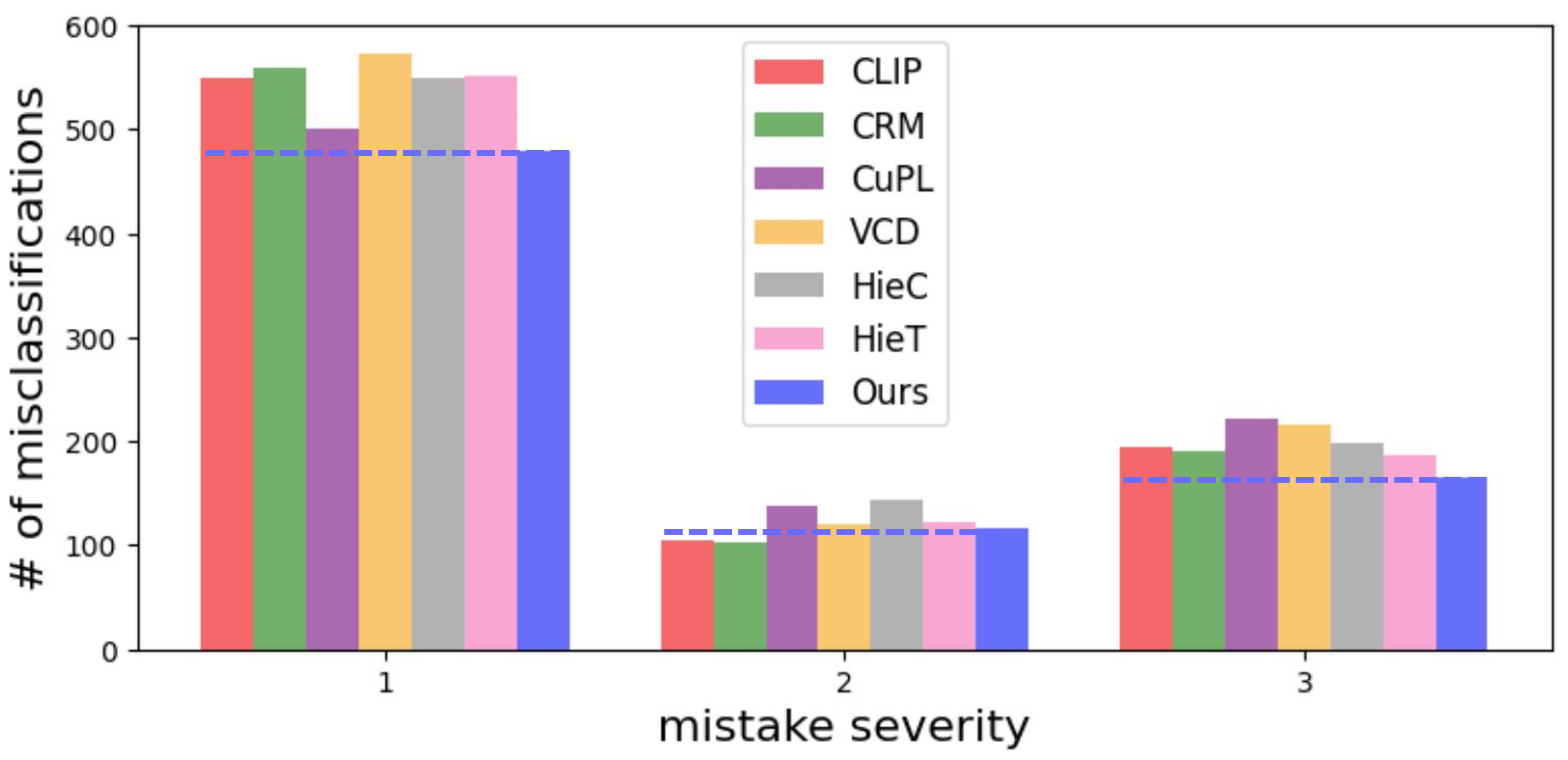} \\
    CUB-200  & SUN-324 \\
    \includegraphics[height=1.2in]{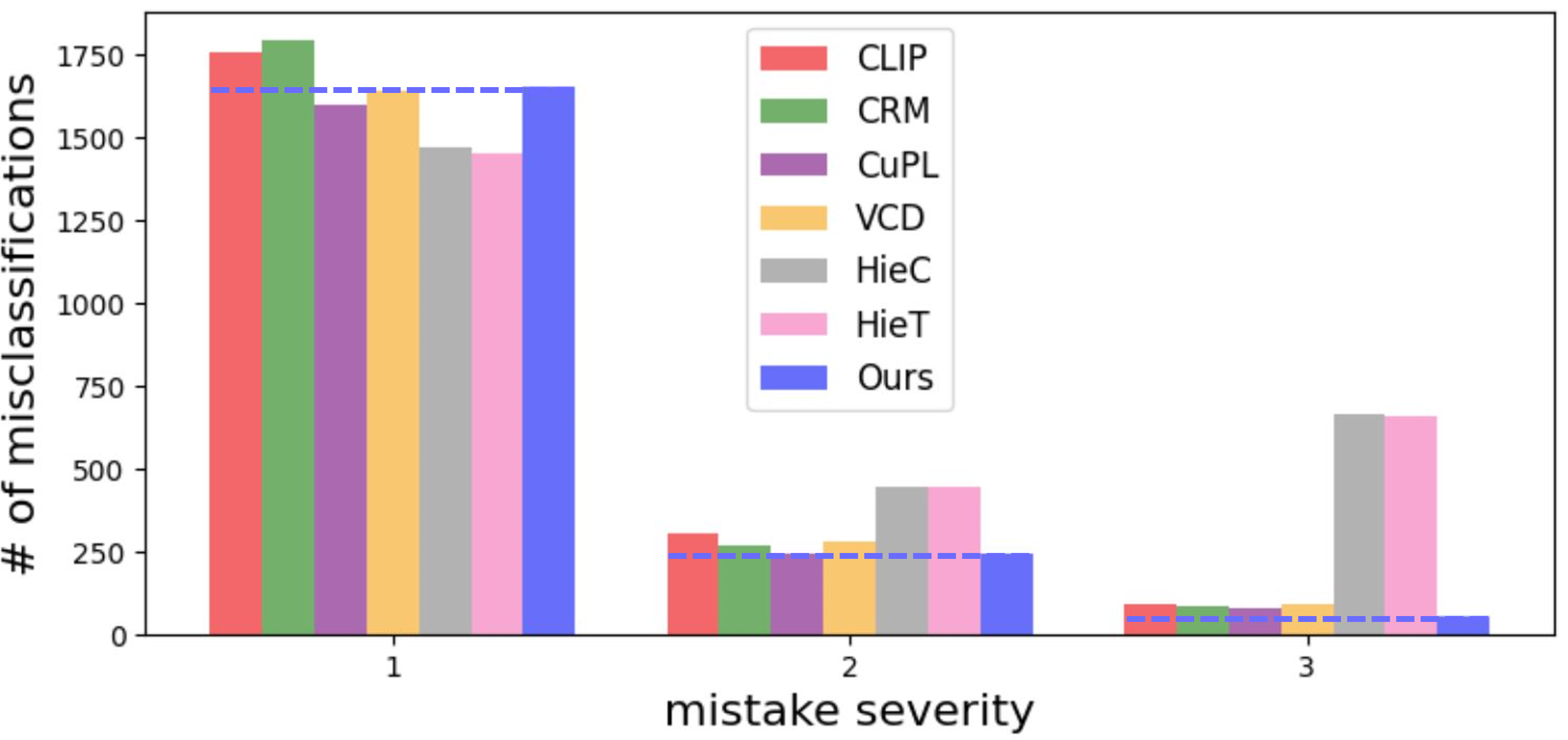} &   \includegraphics[height=1.2in]{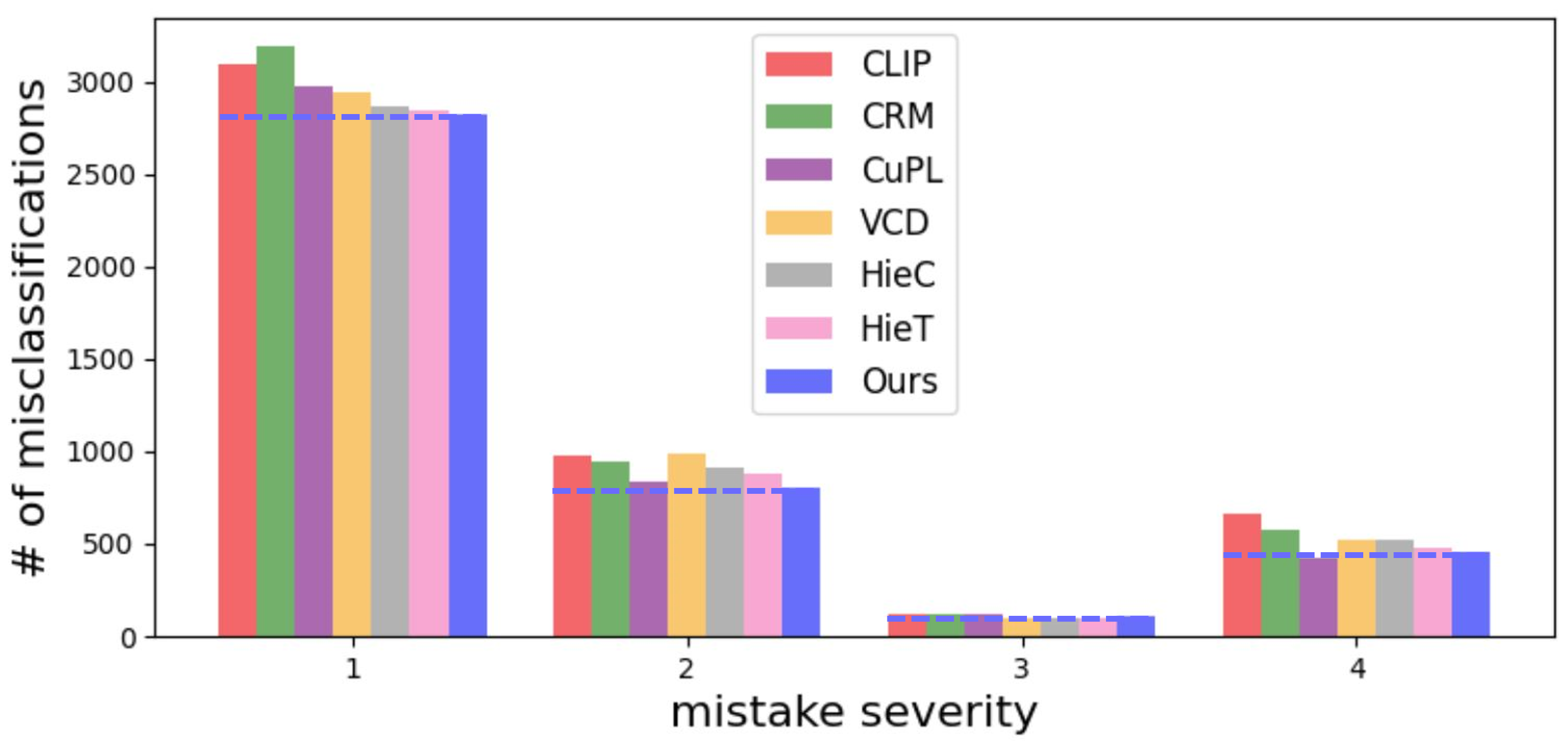} \\
    \multicolumn{2}{c}{ImageNet-1k} \\
    \multicolumn{2}{c}{\includegraphics[height=1.4in]{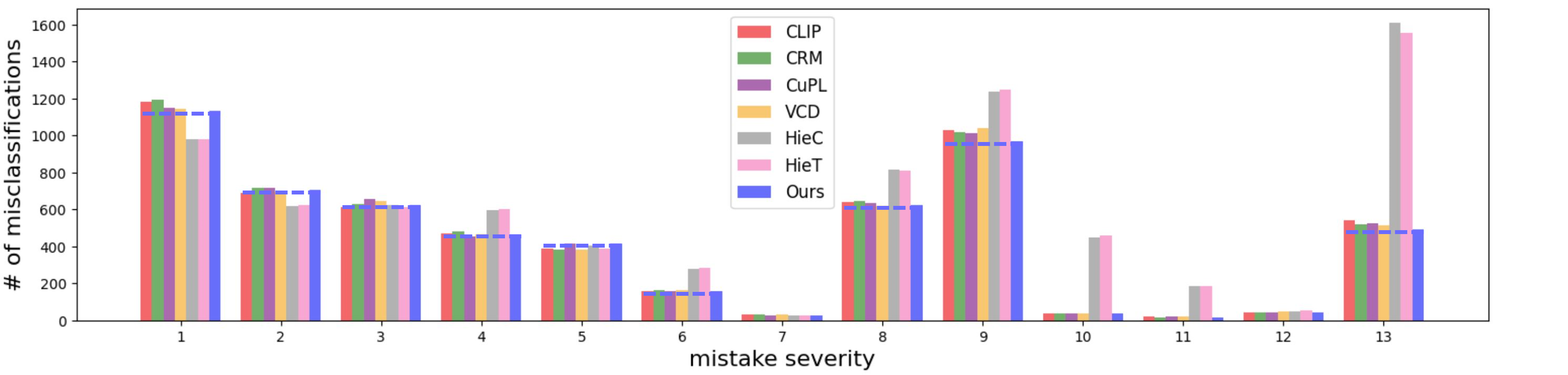}} \\
    \end{tabular}
    \caption{Histograms of mistake severities of all methods' predictions on the five datasets employed in our experiment. We add horizontal dashed lines at the height of our approach's histogram bars when applicable for better visual clarity.}
    \label{fig:histogram_all}
\end{figure*}

\subsection{Histogram of Mistake Severities for Each Dataset} \label{sect:histogram_full}
In this section, we show histograms of mistake severities for the test set predictions of all methods on Food-101, UCF-101, CUB-200, SUN-324, and ImageNet-1k in Fig.~\ref{fig:histogram_all}. In SUN-324, our approach made slightly more high-severity mistakes (severity$=4$) than CuPL. For all other datasets, our approach made fewer high-severity mistakes (i.e., when severity is over half of the maximum possible severity) than other approaches.

\subsection{Visual Examples with HAPrompts} \label{sect:visual_example}

\begin{figure*}
    \centering
    \scriptsize
    \setlength{\tabcolsep}{8pt}
    \begin{tabular}{c| c| l | l}
    \hline
    Dataset & Test Image & Predictions (Severities) & Example Image Prompt of Ground Truth (ours) \\
    \hline
         Food-101 &
         \makecell[tc]{\includegraphics[height=0.8in]{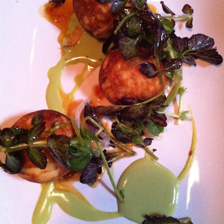}} &  
         \makecell[bl]{\textcolor{blue}{Ground truth: Takoyaki (0)}\\ CLIP: Crab Cakes (4)\\CRM: Crab Cakes (4)\\CuPL: Gnocchi (4)\\VCD: Crab Cakes (4)\\HieC: Crab Cakes (4)\\HieT: Crab Cakes (4)\\ \textcolor{blue}{HAPrompts: Takoyaki (0)}} & 
         \makecell[bl]{``Takoyaki is a Japanese street food that consists \\of savory dough balls filled with octopus pieces,\\ tempura scraps, pickled ginger, and green onion, \\topped with takoyaki sauce and bonito flakes''} \\
    \hline
    \hline 
         UCF-101 &
         \makecell[tc]{\includegraphics[height=0.8in]{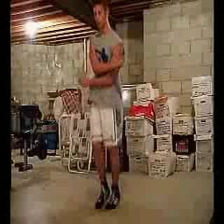}} & 
         \makecell[bl]{\textcolor{blue}{Ground truth: Jump Rope (0)}\\CLIP: Boxing Punching Bag (3)\\CRM: Boxing Punching Bag (3)\\CuPL: Boxing Punching Bag (3)\\VCD: Wall Pushups (3)\\HieC: Wall Pushups (3)\\HieT: Wall Pushups (3)\\ \textcolor{blue}{HAPrompts: Jump Rope (0)}} & 
         \makecell[bl]{``Jump rope is a form of physical exercise that\\ involves jumping over a rope that is swung over \\the body, usually in rhythmic patterns''} \\
    \hline     
    \hline 
        CUB-200 & 
        \makecell[tc]{\includegraphics[height=0.8in]{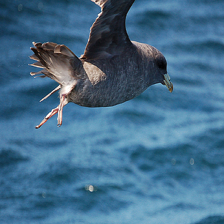}} & 
        \makecell[bl]{\textcolor{blue}{Ground truth: Northern Fulmar (0)}\\CLIP: Least Auklet (3)\\CRM: Least Auklet (3)\\CuPL: Pomarine Jaeger (3)\\VCD: Least Auklet (3)\\HieC: Le Conte Sparrow (3)\\HieT: Pomarine Jaeger (3)\\ \textcolor{blue}{HAPrompts: Northern Fulmar (0)}} &
        \makecell[bl]{``Northern fulmars are smaller and stockier birds \\with a wingspan of around 39 inches, while \\Diomedeidae (albatrosses) are much larger with\\ wingspans that can reach up to 11 feet''} \\
    \hline
    \hline
        SUN-324 & 
        \makecell[tc]{\includegraphics[height=0.8in]{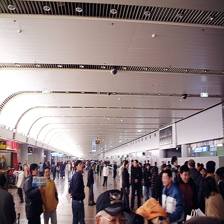}} & 
        \makecell[bl]{\textcolor{blue}{Ground truth: Airport Terminal (0)}\\CLIP: Arrival Gate Outdoor (4)\\CRM: Arrival Gate Outdoor (4)\\CuPL: Arrival Gate Outdoor (4)\\VCD: Arrival Gate Outdoor (4)\\HieC: Arrival Gate Outdoor (4)\\HieT: Arrival Gate Outdoor (4)\\ \textcolor{blue}{HAPrompts: Airport Terminal (0)}} &
        \makecell[bl]{``Airport terminals are typically larger, open\\ spaces with check-in counters, security \\checkpoints, boarding gates, shops, restaurants, \\and other amenities for travelers''} \\
    \hline
    \hline
        ImageNet & 
        \makecell[tc]{\includegraphics[height=0.8in]{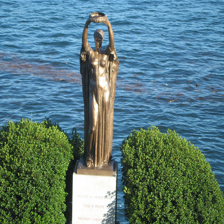}} & 
        \makecell[bl]{\textcolor{blue}{Ground truth: Lakeshore (0)}\\CLIP: Fountain (13)\\CRM: Fountain (13)\\CuPL: Fountain (13)\\VCD: Fountain (13)\\HieC: Fountain (13)\\HieT: Fountain (13)\\ \textcolor{blue}{HAPrompts: Lakeshore (0)}} & 
        \makecell[bl]{``Lakeshores typically have a flatter, more \\open landscape compared to valleys which\\ are often surrounded by steep, sloping hills \\or mountains''} \\
        
    \hline
    \end{tabular}
    \caption{Our HAPrompts' correct prediction examples and our associated sample image prompt. We provide our complete sets of language and image prompts in our GitHub repo. All test images are resized to $224\times 244$ for visual clarity. The numbers inside the parentheses after the predicted label are the associated mistake severities.}
    \label{fig:positive_examples}
\end{figure*}

\begin{figure*}[h]
    \centering
    \scriptsize
    \setlength{\tabcolsep}{8pt}
    \begin{tabular}{c| c| l | l}
    \hline
    Dataset & Test Image & Predictions (Severities) & Example Image Prompt of Our Predicted Class \\
    \hline
         Food-101 &
         \makecell[tc]{\includegraphics[height=0.8in]{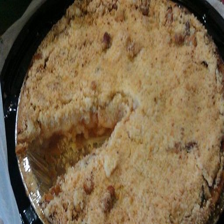}} &  
         \makecell[bl]{\textcolor{blue}{Ground truth: Apple Pie (0)}\\ CLIP: Apple Pie (0)\\CRM: Apple Pie (0)\\CuPL: Lasagna (4)\\VCD: Lasagna (4)\\HieC: Lasagna (4)\\HieT: Lasagna (4)\\ \textcolor{red}{HAPrompts: Carrot cake (1)}} & 
         \makecell[bl]{``Carrot cake and apple pie have distinct appearances\\ that make them easy to differentiate: Carrot cake is a\\ dense, moist cake with a rich orange color from the\\ shredded carrots''} \\
    \hline
    \hline 
         UCF-101 &
         \makecell[tc]{\includegraphics[height=0.8in]{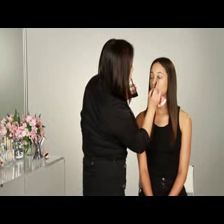}} & 
         \makecell[bl]{\textcolor{blue}{Ground truth: Apply Eye Makeup (0)}\\CLIP: Apply Eye Makeup (0)\\CRM: Apply Eye Makeup (0)\\CuPL: Apply Lipstick (1)\\VCD: Apply Lipstick (1)\\HieC: Apply Lipstick (1)\\HieT: Apply Eye Makeup (0)\\ \textcolor{red}{HAPrompts: Apply Lipstick (1)}} & 
         \makecell[bl]{``Applying lipstick involves using a lip brush or \\directly applying the lipstick onto the lips''} \\
    \hline     
    \hline 
        CUB-200 & 
        \makecell[tc]{\includegraphics[height=0.8in]{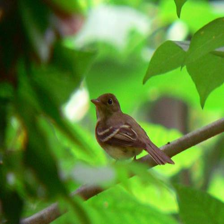}} & 
        \makecell[bl]{\textcolor{blue}{Ground truth: Acadian Flycatcher (0)}\\CLIP: Acadian Flycatcher (0)\\CRM: Acadian Flycatcher (0)\\CuPL: Least Flycatcher (1)\\VCD: Acadian Flycatcher (0)\\HieC: Acadian Flycatcher (0)\\HieT: Acadian Flycatcher (0)\\ \textcolor{red}{HAPrompts: ovenbird (2)}} &
        \makecell[bl]{``The ovenbird is a medium-sized bird with a \\brownish-olive back, white underparts with dark \\spots, a boldly striped crown, and a distinctive \\orange stripe on its head''} \\
    \hline
    \hline
        SUN-324 & 
        \makecell[tc]{\includegraphics[height=0.8in]{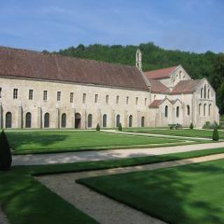}} & 
        \makecell[bl]{\textcolor{blue}{Ground truth: Abbey (0)}\\CLIP: Abbey (0)\\CRM: Abbey (0)\\CuPL: Monastery Outdoor (1)\\VCD: Abbey (0)\\HieC: Abbey (0)\\HieT: Abbey (0)\\ \textcolor{red}{HAPrompts: Monastery Outdoor (1)}} &
        \makecell[bl]{``A monastery typically has a more secluded and\\ peaceful outdoor setting, with gardens, courtyards, \\and areas for contemplation and meditation''} \\
    \hline
    \hline
        ImageNet & 
        \makecell[tc]{\includegraphics[height=0.8in]{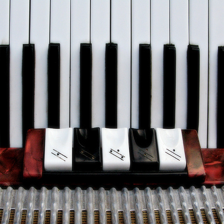}} & 
        \makecell[bl]{\textcolor{blue}{Ground truth: Accordion (0)}\\CLIP: Upright Piano (5)\\CRM: Upright Piano (5)\\CuPL: Upright Piano (5)\\VCD: Grand Piano (5)\\HieC: Upright Piano (5)\\HieT: Grand Piano (5)\\ \textcolor{red}{HAPrompts: Upright Piano (5)}} & 
        \makecell[bl]{``An upright piano typically has a large wooden body\\ with a keyboard, while a percussion instrument may\\ be made of various materials like metal, wood, or \\plastic and does not have a keyboard''} \\
        
    \hline
    \end{tabular}
    \caption{Our HAPrompts' incorrect prediction examples and our associated sample image prompt. We provide our complete sets of language and image prompts in our GitHub repo. All test images are resized to $224\times 244$ for visual clarity. The numbers inside the parentheses after the predicted label are the associated mistake severities.}
    \label{fig:negative_examples}
\end{figure*}

We provide visual examples of our approaches' correct and incorrect predictions with one respective instance of our predicted class' image prompts generated by LLM in Fig.~\ref{fig:positive_examples} and Fig.~\ref{fig:negative_examples}, respectively. In Fig.~\ref{fig:positive_examples}, one can see that our HAPrompts may list a series of objects that are likely to co-occur with the query class: \textcolor{blue}{``\textbf{Takoyaki} is a Japanese street food that consists of \textbf{savory dough balls} filled with \textbf{octopus pieces}, \textbf{tempura scraps}, \textbf{pickled ginger}, and \textbf{green onion}, topped with takoyaki sauce and \textbf{bonito flakes}''}, for query class ``Takoyaki''. This list of likely co-occurring objects could help improve classification accuracy as it may result in the associated class text embedding being closer to the test image embeddings. A side effect of our comparative prompts is also shown in both Fig.~\ref{fig:positive_examples} and Fig.~\ref{fig:negative_examples}, the associated image prompts generated by LLM could include both the query class and the related class in our comparative language prompt: \textcolor{blue}{``\textbf{Northern fulmars} are smaller and stockier birds with a wingspan of around 39 inches, while \textbf{Diomedeidae (albatrosses)} are much larger with wingspans that can reach up to 11 feet''} for query class ``Northern Fulmars'', and \textcolor{blue}{``An \textbf{upright piano} typically has a large wooden body with a keyboard, while a \textbf{percussion instrument} may be made of various materials like metal, wood, or plastic and does not have a keyboard''} for query class ``Upright Piano''. We provide our complete sets of language and image prompts in our GitHub repo at \href{https://github.com/ltong1130ztr/HAPrompts/tree/master/lang_prompts}{language prompts link} and \href{https://github.com/ltong1130ztr/HAPrompts/tree/master/image_prompts}{image prompts link}. 

\end{document}